\documentclass[10pt,a4paper]{article}

\usepackage[margin=1in]{geometry}
\usepackage{setspace}
\usepackage{newtxtext,newtxmath}  
\usepackage{amsmath}
\usepackage{graphicx}
\usepackage{amssymb}
\usepackage{titlesec}
\usepackage[numbers,sort&compress]{natbib}  

\let\oldcite\cite
\renewcommand{\cite}[1]{\textbf{\oldcite{#1}}}

\usepackage{float}
\usepackage{indentfirst}
\usepackage{makecell}
\usepackage{array}

\usepackage{siunitx}
\usepackage[linesnumbered,ruled,vlined]{algorithm2e}
\usepackage{tabularx}
\usepackage{booktabs}
\usepackage[font=footnotesize, labelfont=bf]{caption} 
\usepackage{subcaption}
\usepackage{wrapfig}
\usepackage[hidelinks]{hyperref}
\usepackage{svg}
\usepackage{fancyhdr}

\title{A Behaviour-Aware Federated Forecasting Framework for Distributed Stand-Alone Wind Turbines\thanks{Code available at \url{https://github.com/bowl1/Wind-and-AI}}}

\author{
Bowen Li\thanks{IT University of Copenhagen, Denmark. Email: bowenivy0@gmail.com} \and
Xiufeng Liu\thanks{Technical University of Denmark (DTU), Denmark} \and
Maria Sinziiana Astefanoaei\footnotemark[2]
}

\date{} 

\begin{document}
\maketitle

\begin{abstract}
Accurate short-term wind power forecasting is essential for grid dispatch and market operations, 
yet centralising turbine data raises privacy, cost, and heterogeneity concerns.
We propose a two-stage federated learning framework that first clusters turbines by long-term behavioural statistics 
using a novel Double Roulette Selection (DRS) initialisation with recursive Auto-split refinement, then trains cluster-specific LSTM models via FedAvg.
On a real-world dataset of 400 stand-alone turbines in Denmark, our DRS-auto method discovers seven behaviourally coherent clusters 
and achieves competitive forecasting accuracy  while preserving data locality. 
Experiments show that behaviour-aware grouping outperforms geographic partitioning and matches centralised K-means++ baselines, 
demonstrating a practical, privacy-preserving solution for distributed turbine fleets.
\end{abstract}


\pagestyle{plain}     
\pagenumbering{arabic}

\section{Introduction}
\label{sec:introduction}

Wind power is becoming a major component of modern energy systems, yet its intrinsic
variability and intermittency make accurate short-term forecasting essential for grid
dispatch and market operations~\cite{li2023wind}. Many learning-based forecasting
pipelines rely on centralising turbine or wind-farm data for model training, which
raises practical concerns: operational data can be commercially sensitive, uploading
large-scale time series can be costly, and strong heterogeneity across turbines and
sites can degrade the performance of a single global model under non-IID
conditions~\cite{cheng2022review,li2023wind}. Federated learning (FL) mitigates these
issues by keeping raw data local and aggregating model updates instead of time
series~\cite{mcmahan2017communication}. However, existing FL work in wind forecasting
often treats each wind farm as a homogeneous client, which is inadequate for fleets of
independent turbines whose behaviour differs substantially in operating regimes,
availability, and control strategies.

To address turbine-level heterogeneity under privacy constraints, we propose a two-stage
framework that first discovers behaviourally similar turbine groups via federated
clustering, and then trains a dedicated federated forecasting model within each group.
Unlike location-based partitioning, our grouping is driven by long-term behavioural
statistics extracted locally from turbine power time series, enabling functional
similarity while avoiding raw-data centralisation.

\paragraph{Contributions.}
This work makes the following contributions:
\begin{enumerate}
    \item \textbf{Behaviour-aware federated clustering for standalone wind turbines.}
    We introduce a federated K-means clustering pipeline with Double Roulette Selection
    (DRS) initialisation and a recursive Auto-split procedure. Clusters are refined
    using silhouette-based splitting and minimum-size constraints, producing coherent
    behaviour groups using only locally computed summary statistics (e.g., mean power,
    variability, zero-power ratio, and ramping metrics) without sharing raw time series.

    \item \textbf{Cluster-specific federated LSTM forecasting.}
    We treat each behaviour cluster as an independent FL task and train a shared LSTM-based
    short-term forecasting model via FedAvg~\cite{mcmahan2017communication}, reducing
    within-task heterogeneity compared to a single global model.

    \item \textbf{Empirical evaluation with rolling forecasts.}
    We compare the proposed framework against geographic grouping, flat (non-recursive)
    federated K-means baselines, and a centralised LSTM reference, evaluating both
    clustering quality and forecasting accuracy. We further implement 3-step-ahead models
    ($H=3$) and construct 24-hour rolling forecasts from these predictions to visualise
    measured versus forecasted power for representative turbines.
\end{enumerate}
\section{Background and Related Work}
\label{sec:related_work}

\subsection{Federated learning for wind-farm--level power forecasting}
Federated learning (FL)~\cite{mcmahan2017communication} has recently been explored for
wind power forecasting, predominantly at the wind-farm level, where each farm acts as a
client and only model updates are aggregated at the server. Representative approaches
include federated deep sequence models for farm-level forecasting~\cite{ahmadi2022deep},
sample-size--weighted aggregation across multiple farm clients~\cite{alshardan2024federated},
and federated tree ensembles such as FL-XGBoost~\cite{yang2024wind}. Beyond global models,
personalised FL variants have also been investigated to better handle inter-farm
differences~\cite{zhao2024ultra}. Despite these advances, the common modelling granularity
remains coarse: turbines are implicitly assumed to be homogeneous within a farm, and a
single client-level model is used to represent the aggregated farm output. This
assumption becomes inadequate for fleets of highly heterogeneous and independently
operated small turbines.

\subsection{Turbine-level clustering and power forecasting}
At turbine level, many data-driven forecasting methods have been developed under
centralised data access, including classical machine learning and deep learning
approaches surveyed in~\cite{yang2024survey}. Examples include turbine-level short-term
forecasting with random forests~\cite{rashid2020forecasting}, real-time deep models based
on RNN/CNN architectures~\cite{sun2021real}, and LSTM-based turbine predictors that
incorporate operating conditions and train turbine-specific models~\cite{liu2023wind}.
In parallel, clustering has been used to describe turbine heterogeneity and construct
cluster-specific models, e.g., via similarity-based spectral clustering of power
time series~\cite{ma2009cluster} or feature-based Canopy--K-means++ schemes~\cite{zhixuan2023multi}.
However, these turbine-level forecasting and clustering pipelines typically assume
centralised access to raw SCADA or power data, and are not designed for cross-owner
collaboration under privacy constraints.

Existing turbine-level FL studies have primarily focused on condition monitoring and
anomaly detection rather than short-term power forecasting. For instance, FL has been
used to model normal behaviour for fault detection~\cite{jenkel2023privacy} and to train
cross-farm LSTM models for gearbox temperature prediction with client-side fine-tuning to
improve robustness under heterogeneity~\cite{grataloup2024wind}. 

\paragraph{Overall.}prior work either (i) performs turbine-level forecasting/clustering with
centralised data, or (ii) applies FL mainly to monitoring tasks instead of turbine-level
short-term power forecasting. This gap motivates our two-stage framework: we first apply
federated clustering to discover behaviourally similar turbine groups without
centralising raw data, and then train a dedicated federated forecasting model within
each cluster to improve turbine-level prediction under privacy constraints.
\section{Problem Formulation and System Overview}
\subsection{Scenario and client definition}

This work considers a distributed wind power system of 400 independent small-scale turbines in Denmark. Each turbine is treated as a federated learning client that stores time-series data locally (active power, wind speed, direction, temperature) and only shares model updates with the central server. The turbines are spatially distributed across residential, farm, and small-business sites with heterogeneous ownership, rated capacities, control strategies, and operating conditions, resulting in highly non-IID output behaviour.

\begin{figure}[H]
    \centering
    \begin{subfigure}[b]{0.47\textwidth}
        \centering
        \includegraphics[width=\linewidth]{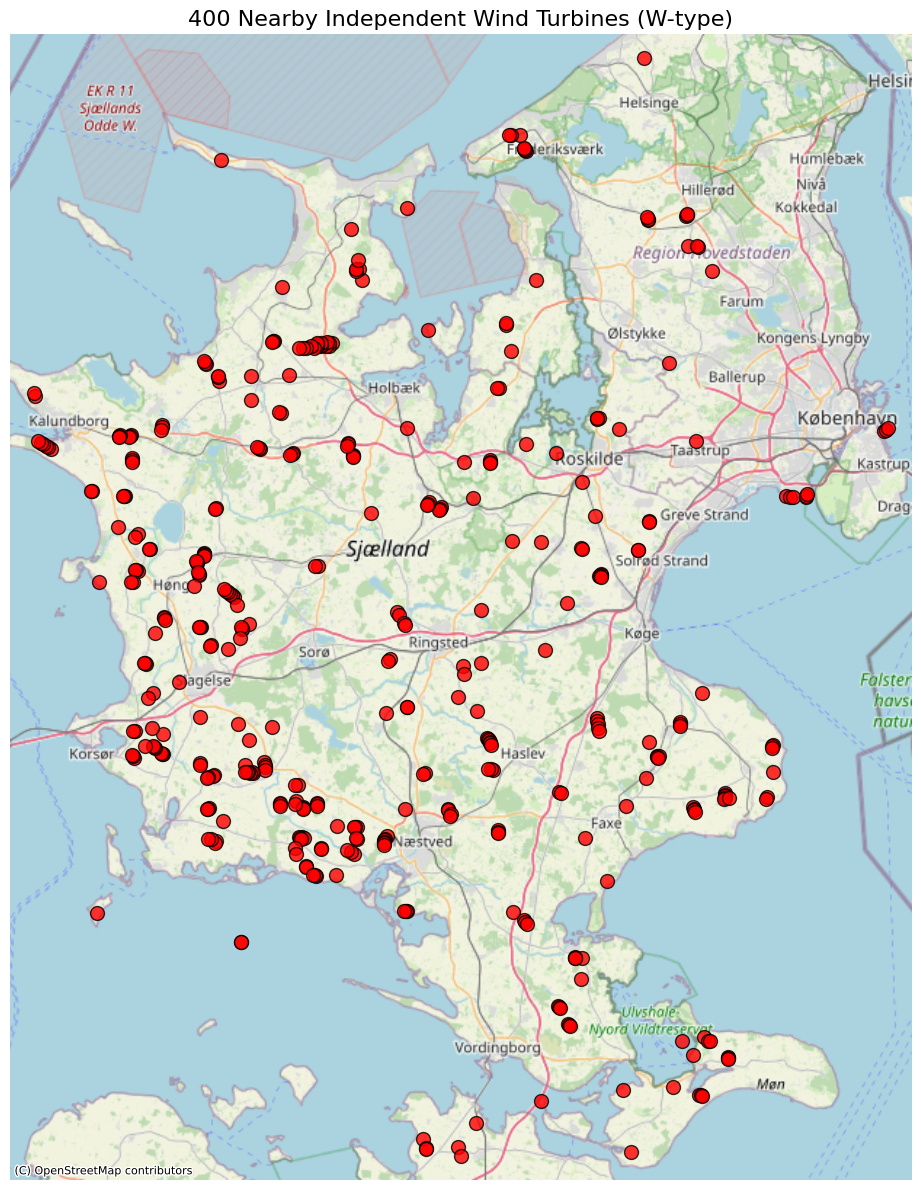}
        \caption{Selected turbines in Denmark}
        \label{fig:mapOverview}
    \end{subfigure}
    \hfill
    \setcounter{subfigure}{2}
    \begin{subfigure}[b]{0.47\textwidth}
        \centering
        \includegraphics[width=\linewidth]{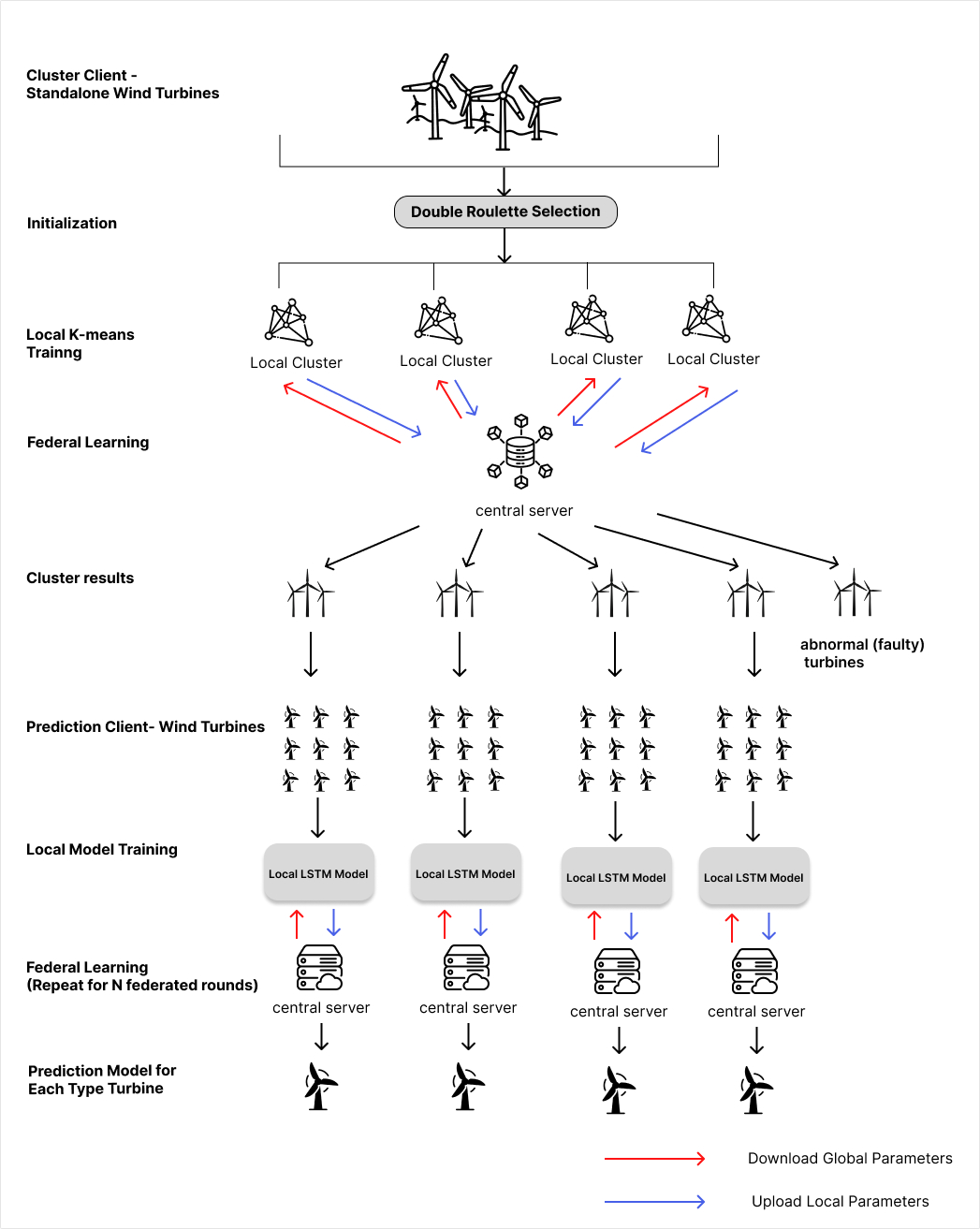}
        \caption{Two-stage federated framework}
        \label{fig:roulette}
    \end{subfigure}

    \vspace{0.6em}

    \setcounter{subfigure}{1}
    \begin{subfigure}[b]{0.75\textwidth}
        \centering
        \includegraphics[width=\linewidth]{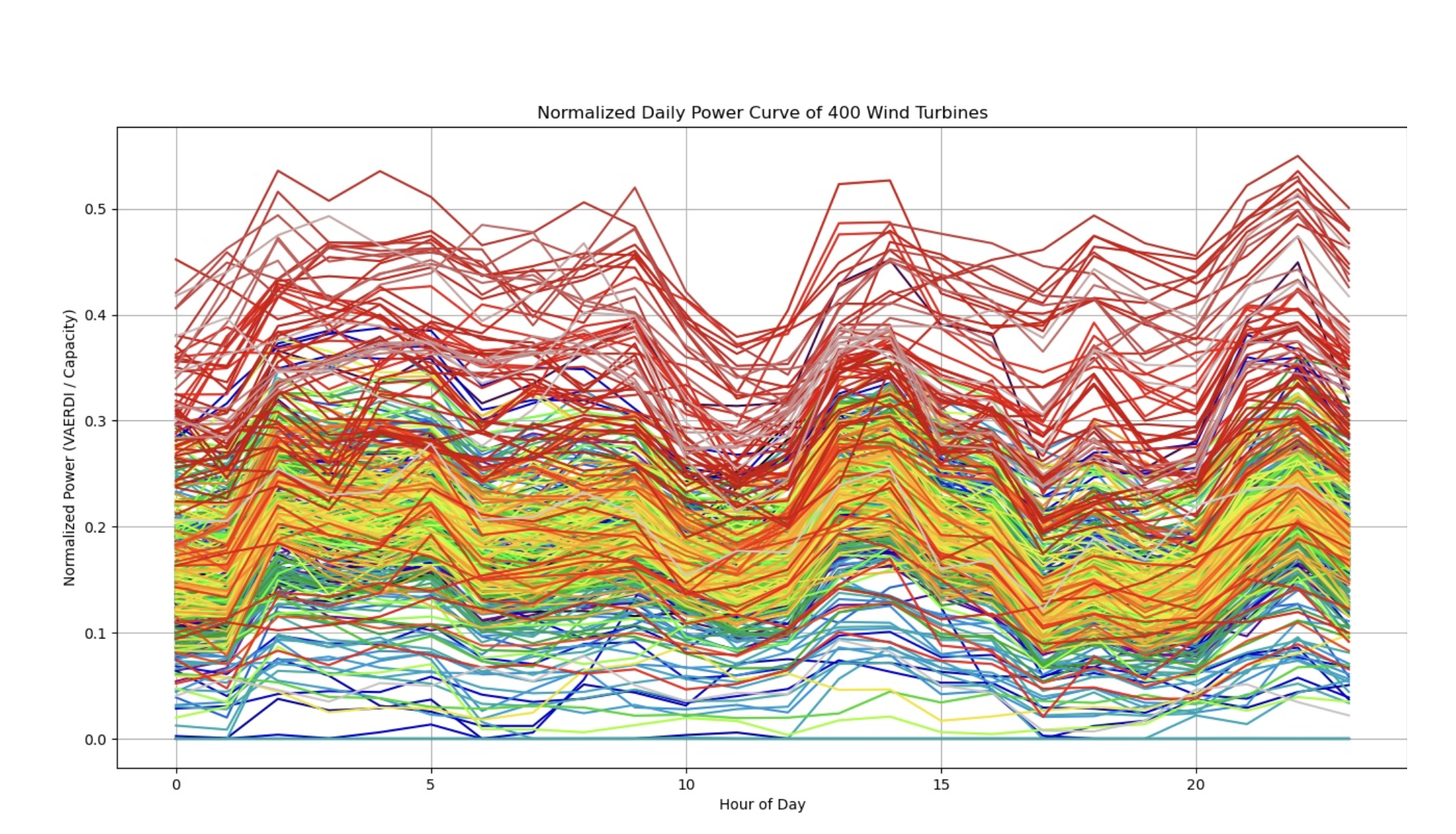}
        \caption{Daily power variation of 400 turbines}
        \label{fig:visualization_31}
    \end{subfigure}
    \caption{(a) 400 turbines selected via nearest-neighbour sampling. (b) Daily generation variation of 400 turbines over one day. (c) Federated clustering followed by cluster-specific federated LSTM forecasting.}
    \label{fig:overview}
\end{figure}

To construct a geographically coherent client set while preserving behavioural diversity, we apply a three-step filtering procedure: (i) retain turbines with positive annual mean generation in 2019, (ii) select standalone turbines with valid coordinates, and (iii) apply nearest-neighbour sampling~\cite{cover1967nearest} to select the 400 spatially closest units (see \textbf{Figure~\ref{fig:mapOverview}}). This design partially decouples geographical from behavioural differences, so that subsequent clustering reflects intrinsic operational patterns rather than location alone.

\subsection{Short-term turbine-level forecasting task}

On the time-series level, we focus on short-term power forecasting at the turbine level. For each turbine, the model observes the past $L = 24$ time steps (1 hour each) of measurements---including active power, wind speed, wind direction (encoded by sine/cosine), temperature, and static turbine features---and aims to predict the \emph{next three} time steps of active power output.

By rolling this three-step-ahead predictor along the time axis and concatenating the overlapping predictions, we construct a 24-hour forecast trajectory for each individual turbine, which can then be used both for evaluation and for downstream decision-making by grid operators and turbine owners.

\subsection{Overview of the two-stage federated framework}

The proposed system adopts a two-stage framework (\textbf{Figure~\ref{fig:roulette}}):
\begin{itemize}
    \item \textbf{Stage~1 (Federated Clustering):} Each turbine locally computes behavioural statistics (mean power, variability, zero-power ratio, ramping metrics) from its historical data. These statistics are used in a federated K-means procedure with Double Roulette Selection (DRS) initialisation and recursive Auto-split refinement to discover behaviourally coherent clusters.
    \item \textbf{Stage~2 (Federated Forecasting):} Each cluster is treated as an independent FL task. Turbines within the same cluster collaboratively train an LSTM model via FedAvg, sharing only model parameters. At inference, each turbine uses its cluster-specific model to produce 3-step-ahead forecasts, which are rolled to construct 24-hour trajectories.
\end{itemize}

\section{Feature Design}

\subsection{Turbine Behaviour Clustering Features}
\label{subsec:turbine_behaviour_features}

In the first, behaviour-driven clustering stage, turbines are grouped according
to long-term operational characteristics rather than purely geographic location.
Since strong variability and rapid changes in wind power output are known to
stress power system operation and reserve requirements~\cite{holttinen2009impact},
explicitly encoding volatility and ramping behaviour is particularly relevant
for identifying distinct turbine operating patterns.

For each turbine, we construct a compact set of statistical features from one year of power time series to characterise its long-term level, variability, availability, and short-term dynamics.

\textbf{Mean power} (\texttt{mean\_power}) is the annual average power output, approximating the long-term capacity factor and resource utilisation level. After standardisation, a positive cluster-level mean ($>0$) indicates turbines that tend to produce above-average power, whereas a negative value ($<0$) corresponds to below-average long-term production.

\textbf{Power standard deviation} (\texttt{std\_power}) measures the amplitude of power fluctuations. In the standardised feature space, a positive cluster mean indicates more volatile output compared to the population average, while a negative mean corresponds to smoother and more stable generation.

\textbf{Coefficient of variation} (\texttt{cv}), defined as $\mathrm{cv} = \mathrm{std\_power}/\mathrm{mean\_power}$, is a standard normalised measure of dispersion~\cite{everitt2002cambridge}. As a dimensionless quantity, it captures relative variability across turbines with different capacities and mean power levels.

\textbf{Zero-power ratio} (\texttt{zero\_ratio}) is the fraction of time steps with zero power output, reflecting turbine availability and shutdown behaviour. Values closer to $1$ indicate frequent or prolonged shutdowns, while values closer to $0$ correspond to high availability.

\textbf{Ramp statistics} (\texttt{ramp\_mean}, \texttt{ramp\_std}) capture short-term dynamics. Power ramps are defined as differences between consecutive time steps. A positive standardised ramp mean indicates a long-term ramping-up trend, whereas a negative value suggests gradual degradation. The ramp standard deviation captures the irregularity and intensity of short-term dynamics.

All features are standardised across the turbine population prior to clustering.
In this standardised feature space, cluster-level means provide directional
interpretability (above- versus below-average behaviour), while the corresponding
within-cluster standard deviations quantify behavioural homogeneity. Clusters
with small standard deviations are behaviourally coherent, whereas clusters with
standard deviations clearly above $1$ indicate substantial internal
heterogeneity.

Taken together, these features form a concise \emph{behaviour fingerprint} for
each turbine, spanning long-term production level, relative variability,
availability, and short-term dynamics. Applying K-means together with the
recursive Auto-split procedure in this feature space groups turbines with
similar operating patterns into the same cluster, while simultaneously
highlighting heterogeneous or abnormal groups. The resulting behaviour clusters
then serve as modelling units for the second-stage federated prediction.

\subsection{Prediction Model Input Features}
\label{subsec:prediction_features}

For short-term forecasting, we construct time-step level inputs following~\cite{costa2008review}. The feature set comprises four groups (\textbf{Table~\ref{tab:pred_features}}). The cluster label is used only for model assignment, not as input.

\begin{table}[ht]
\centering\scriptsize
\caption{Prediction model input features.}
\label{tab:pred_features}
\begin{tabular}{ll}
\toprule
\textbf{Group} & \textbf{Features} \\
\midrule
Meteorological & \texttt{wind\_speed}, \texttt{wind\_dir\_sin/cos}, \texttt{temperature} \\
Temporal & \texttt{hour\_sin/cos} (diurnal encoding) \\
Static & \texttt{Capacity\_kw}, \texttt{age} \\
Autoregressive & \texttt{power\_output\_lag\_1} to \texttt{lag\_24} ($L{=}24$~h window) \\
\bottomrule
\end{tabular}
\end{table}

Wind direction uses sine/cosine encoding to handle circularity; temperature captures air density effects. Static features provide turbine-level context, while the 24-step autoregressive window captures short-term dynamics and local unobserved effects.

\section{Federated Clustering Framework}
\label{sec:clustering_framework}

\subsection{Problem formulation}

Assume there are $N$ independent small turbines. In the clustering stage, each turbine is represented by a $d$-dimensional behavioural feature vector
\[
\mathbf{x}_i \in \mathbb{R}^d, \quad i = 1,\dots,N,
\]

The goal is to partition all turbines into clusters $\{C_1,\dots,C_K\}$ such that turbines within the same cluster exhibit similar long-term behaviour, while inter-cluster differences are as large as possible, without centrally storing the original time series. To this end, we adopt a federated K-Means procedure with local updates and global centroid aggregation, and on top of it we add a Double Roulette Selection (DRS) initialisation and an Auto-split recursive splitting mechanism.

\subsection{Federated K-Means++ with Double Roulette Selection initialization}

\begin{wrapfigure}{r}{0.4\textwidth}
    \centering
    \vspace{-1em}
    \includegraphics[width=0.38\textwidth]{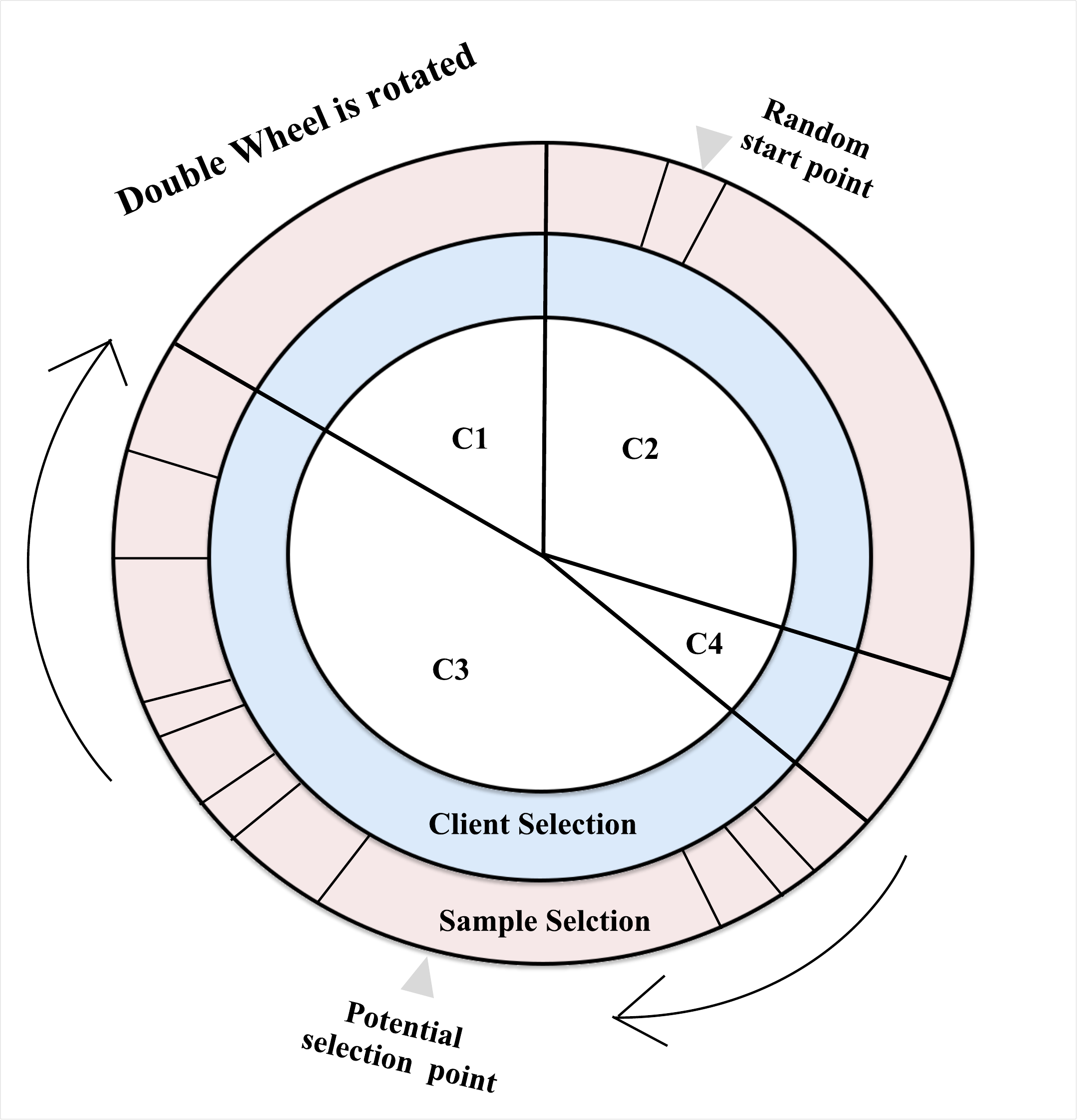} 
    \caption{Double Roulette Selection}
    \label{fig:double_roulette_selection}
    \vspace{-1em}
\end{wrapfigure}

Within the Auto-split framework, whenever a cluster needs further refinement, we run a ``single-round federated K-Means'' on the subset $\mathbf{X}_{\text{sub}} \in \mathbb{R}^{N_{\text{sub}} \times d}$ containing only the turbines currently assigned to that cluster. To emulate a federated setting, $\mathbf{X}_{\text{sub}}$ is randomly partitioned into $n_{\text{clients}}$ disjoint subsets, each subset acting as the local dataset of a logical client.

On top of this, we use \emph{Double Roulette Selection} (DRS) as a global centroid initialisation strategy. To improve the initialization quality of cluster centers under privacy constraints, the second initialization is Double Roulette Selection mechanism, inspired by the k-means++ strategy but adapted for federated settings. The concept of client-level distance sampling draws inspiration from prior work on federated clustering for electricity consumption patterns \cite{wang2022federated}. The goal is to select representative initial centers from distributed clients without requiring full data sharing. \textbf{Figure~\ref{fig:double_roulette_selection}} shows the basic idea of Double Roulette Selection.

\paragraph{(1) Initialising the first centre.}
We randomly pick one sample from all clients and their local data as the first global centre $\mathbf{c}_1$.

\paragraph{(2) Two-level roulette for the remaining centres.}
For each subsequent centre $\mathbf{c}_m$ ($m = 2,\dots,K$), we perform a two-level roulette sampling. At the \emph{client level}, for each client $p$, we compute the total squared distance of its local samples to the current set of centres $C = \{\mathbf{c}_1,\dots,\mathbf{c}_{m-1}\}$:
\[
D_p = \sum_{i=1}^{N_p} d^2(\mathbf{x}_{p,i}, C),
\]
where $d^2(\mathbf{x}_{p,i}, C)$ denotes the squared distance from $\mathbf{x}_{p,i}$ to its nearest centre in $C$. We then sample a client according to $P(p) = D_p / \sum_q D_q$, so that clients whose data are overall farther from the existing centres are more likely to contribute a new centre. At the \emph{within-client level}, for the selected client $p$, we compute for each local sample its minimum squared distance to $C$, and sample a point according to
\[
P(\mathbf{x}_{p,i} \mid p) = \frac{d^2(\mathbf{x}_{p,i}, C)}{\sum_j d^2(\mathbf{x}_{p,j}, C)}.
\]
The selected sample is added as the next global centre $\mathbf{c}_m$.

This two-level procedure is repeated until $k_{\text{global}}$ initial global centres are obtained. Compared with centralised k-means++, DRS only exposes aggregated distance statistics and a small number of selected samples, achieving good initialisation quality in a more privacy-friendly manner.

\paragraph{(3) Federated local updates and global aggregation.}

Given the initial centres, we perform a few communication rounds of federated K-Means~\cite{garst2024federated}. In the \emph{local update} step, the server broadcasts the current global centres $C^{(t)}$ to all clients. Each client runs one local K-Means iteration: it assigns its local samples to the nearest centre and updates the local centroids $\{\mathbf{c}_{j,k}\}$ as local means, while also counting the number of samples per local cluster $\{S_{j,k}\}$. In the \emph{global aggregation} step, the server collects local centroids and sample counts, and for each cluster index $k$ computes a weighted average:
\[
\mathbf{c}_k^{\text{new}} = \frac{\sum_j S_{j,k}\,\mathbf{c}_{j,k}}{\sum_j S_{j,k}}.
\]

This procedure is analogous to FedAvg~\cite{mcmahan2017communication}, except that we average cluster centres instead of model parameters. After $c_{\text{rounds}}$ communication rounds, we obtain a stable set of global centroids. All samples in the current node are then assigned to the nearest centroid, forming the sub-clusters at this level.

\subsection{Auto-split: silhouette-based recursive splitting}

A single federated K-Means run produces one partition for a given choice of $(n_{\text{clients}}, k_{\text{global}}, c_{\text{rounds}})$. To automatically find a cluster structure that is neither overly fragmented nor overly merged, we propose an Auto-split recursion that, for each node, jointly searches over federated hyperparameters and the number of clusters and uses the silhouette score \cite{rousseeuw1987silhouettes}to decide whether to split further.

\paragraph{(1) Cluster tree structure.}
Auto-split organises the full clustering process as a tree. The root node contains all turbine samples. For any node (cluster) under consideration, we run federated K-Means with multiple candidate configurations. If the node meets the splitting criteria, it is partitioned into several child clusters, which become child nodes in the tree. Nodes that are no longer split are treated as leaf clusters, forming the final behavioural groups.

\paragraph{(2) Parameter grid and evaluation.}

For each cluster to be split, Auto-split searches over a predefined grid
\[
(n_{\text{clients}}, k_{\text{global}}, c_{\text{rounds}}) \in \mathcal{G}.
\]
For each combination in $\mathcal{G}$, we run a single-round federated K-Means to obtain cluster labels $\hat{z}$, and compute the corresponding silhouette score
\[
s = \text{silhouette}(\mathbf{X}_{\text{sub}}, \hat{z}).
\]
We then select the configuration $(n^\ast, k^\ast, c^\ast)$ and partition $\hat{z}^\ast$ that achieve the highest silhouette score $s^\ast$.

\paragraph{(3) Splitting decision and stopping conditions.}

To avoid overfitting and severely imbalanced cluster sizes, Auto-split applies the following rules to each node. Let $|C|$ be the size of the current cluster and $N$ the total number of samples. If $|C|/N$ is below a minimum cluster ratio $\alpha_{\min}$ (implemented as \texttt{min\_ratio}), the cluster is considered small or anomalous and will not be split further. Otherwise, if the best silhouette score $s^\ast$ exceeds a threshold $s_{\min}$ (implemented as \texttt{min\_silhouette}), we assume that the cluster contains well-separated sub-structure and accept the split defined by $\hat{z}^\ast$. If $s^\ast < s_{\min}$ but the cluster is too large (e.g.\ $|C|/N > \alpha_{\max}$, implemented as \texttt{max\_single\_cluster\_ratio}), we allow a one-time ``forced split'' to prevent a single giant cluster from dominating. In all other cases, the node is not split further and becomes a leaf node.

For each split node, the selected hyperparameters and its silhouette score are recorded for later analysis and reporting.

\paragraph{(4) Leaf clusters and final grouping.}

Auto-split recursively repeats the above process until no node satisfies the splitting conditions. The leaves of the resulting tree form the final set of behavioural clusters. Each turbine obtains a unique leaf-cluster label, which is then used to (i) visualise the clusters in the behavioural feature space via PCA or other dimensionality reduction methods; (ii) compute per-cluster statistics (means and standard deviations) of the behavioural features, enabling semantic descriptions such as ``high-stable-power clusters'', ``high-variability clusters'', or ``high-zero-ratio clusters''; and (iii) define the groups for cluster-wise federated LSTM forecasting, where all turbines in the same leaf cluster jointly train a cluster-specific LSTM model.

\section{Federated LSTM-based Forecasting}
\label{sec:fl_forecasting}

\subsection{Problem formulation}
\label{subsec:problem_formulation}

After the federated clustering stage, each turbine is assigned a behavioural cluster
label. We formulate short-term forecasting as a multi-step time-series regression
problem: given a historical window of length $L$ (e.g., $L=24$ time steps), predict the
next $H$ power outputs, where $H=3$ in our experiments. Training is conducted in a
federated setting: each turbine keeps its raw time series locally and only shares model
updates with the server.

\subsection{Local model architecture}
\label{subsec:lstm_architecture}

\begin{figure}[h]
    \centering
    \includegraphics[width=0.55\textwidth]{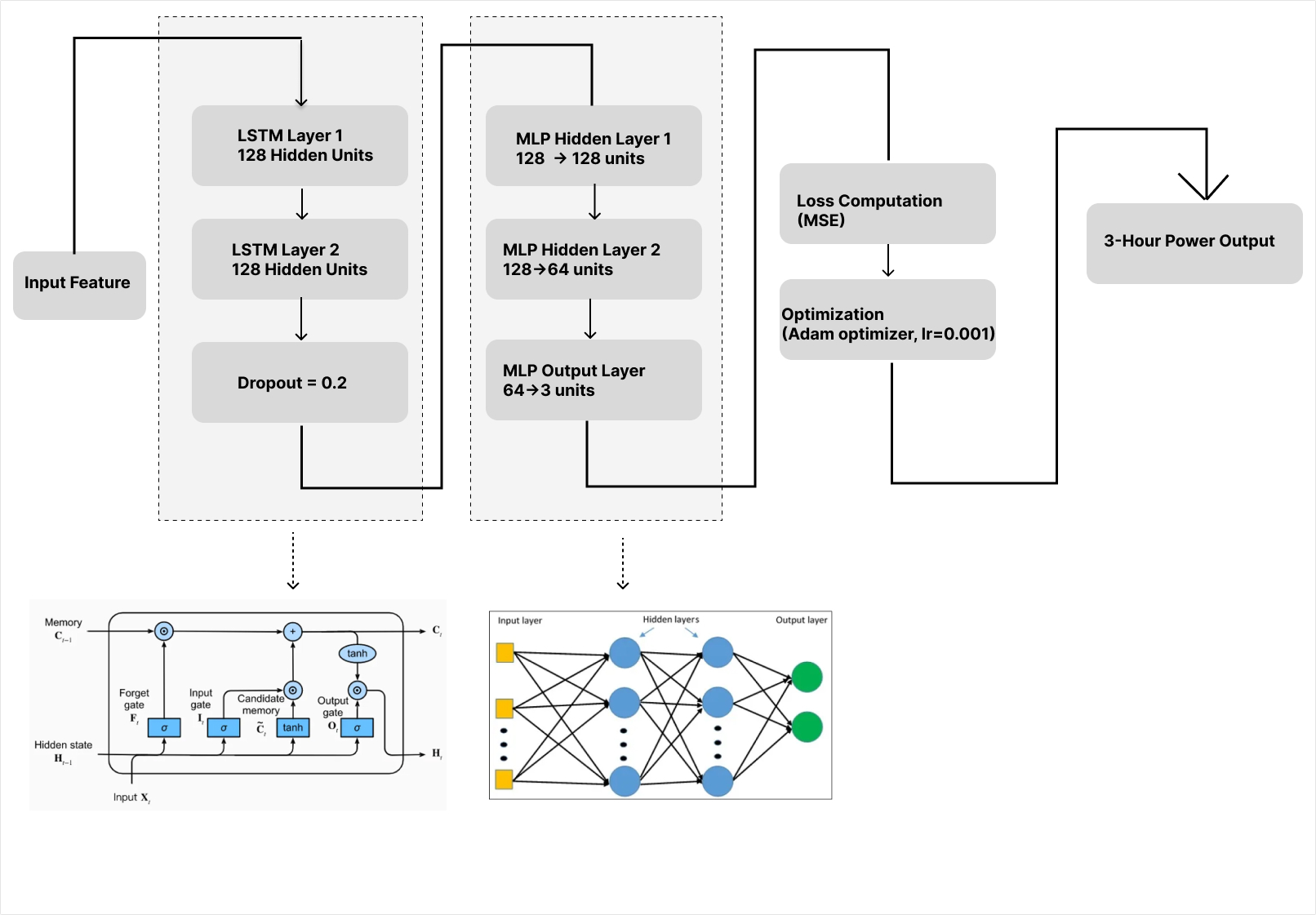}
    \caption{LSTM--MLP architecture.}
    \label{fig:lstm_mlp}
\end{figure}

We adopt an LSTM-based regressor, as LSTM models are a widely used and competitive baseline for wind power time-series forecasting~\cite{hochreiter1997long,yang2024survey}. Each turbine trains the same hybrid LSTM--MLP architecture locally (\textbf{Figure~\ref{fig:lstm_mlp}}). The input sequence  is first encoded by an LSTM to capture temporal dependencies. The final hidden representation is then passed to a lightweight MLP regressor that outputs an $H$-dimensional prediction vector. While we report a 3-step-ahead objective ($H=3$), these forecasts can be used in a rolling manner to construct longer-horizon trajectories (e.g., 24-hour rolling forecasts) when needed.

\subsection{Cluster-wise federated training protocol}
\label{subsec:clusterwise_fedavg}

In the forecasting stage, we run a separate federated learning task for each behavioural cluster (per-cluster FL). For cluster $k$, turbines in $\mathcal{C}_k$ act as clients jointly training a shared forecasting model. We use FedAvg~\cite{mcmahan2017communication} for parameter aggregation in all clusters. Training one model per cluster allows turbines with similar long-term behaviour to share a dedicated predictor, improving statistical efficiency while avoiding the limitations of fitting a single global model to highly heterogeneous turbines under strict privacy
constraints.

\section{Experimental Setup}

\subsection{Experimental Environment}
Experiments are conducted on a local macOS platform with Python~3.10. lustering and visualisation use \texttt{scikit-learn} (KMeans, silhouette, PCA) and \texttt{matplotlib}. Federated training is implemented in PyTorch with the Flower framework~\cite{beutel2020flower}. All random seeds are fixed (\texttt{SEED=42}) for reproducibility. A total of 400 turbines in Denmark are selected via nearest-neighbour sampling.

\begin{table}[t]
\centering\scriptsize
\begin{tabular}{l r}
\toprule
\textbf{Cluster} & \textbf{Training time (min)} \\
\midrule
Cluster 0 & 27 \\
Cluster 1 & 43 \\
Cluster 2 & 3 \\
Cluster 4 & 45 \\
Cluster 5 & 5 \\
Cluster 6 & 7 \\
\midrule
\textbf{Total} & \textbf{130} \\
\bottomrule
\end{tabular}
\caption{Wall-clock training time in the prediction stage (per-cluster FedAvg LSTM).}
\label{tab:pred_time_per_cluster}
\end{table}

\textbf{Table~\ref{tab:pred_time_per_cluster}} reports wall-clock time for the per-cluster federated LSTM training.
Since clusters are independent FL tasks, training can be parallelised; in practice, wall-clock time is dominated by the largest/slowest cluster.

\subsection{Evaluation Metrics}

We use standard regression metrics: MAE, MSE, RMSE, and $R^2$~\cite{bishop2006pattern}. Lower MSE/RMSE/MAE and higher $R^2$ indicate better performance.

\subsection{Auto-split configuration used in the experiments}
\label{subsec:auto_split_config}

In all experiments using DRS-style recursive clustering, the decision of whether a node (as mentioned in section 5.3 Auto-split organises the full clustering process as a tree) should be further split is governed by the following rules. A \emph{silhouette-based split} is performed if the best silhouette score on the current node satisfies $\text{silhouette}_{\text{best}} \geq 0.45$, indicating a meaningful partition. A \emph{forced split} is applied if the current cluster contains more than $70\%$ of all samples ($|C_{\text{node}}|/|X| > 0.7$), even when the silhouette score is relatively low, in order to avoid a single giant cluster dominating most turbines. Conversely, if the node size is at most $30\%$ of the full dataset ($|C_{\text{node}}|/|X| \leq 0.3$), it is regarded as a small terminal cluster or potential outlier group and is directly marked as a leaf, without further recursive splitting.

The auto-split procedure first runs federated KMeans over the parameter grid
\[
n \in [3, 9], \quad
k \in [3, 10], \quad
c \in [3, 10],
\]
where $n$ is the number of clients, $k$ the number of global clusters, and $c$
the number of communication rounds.

For nodes that satisfy the splitting condition, their samples are partitioned into several
child clusters according to the best label assignment obtained from federated KMeans.
A child node is created for each label and pushed into the queue; child clusters whose size
falls below the minimum ratio threshold are instead marked as outlier leaves and are not
expanded further. The algorithm iterates until the queue becomes empty, resulting in a
multi-level clustering tree.

\subsection{Baseline grouping strategies}
\label{subsec:baselines}

To isolate the effect of behaviour-aware recursive clustering, we compare DRS-auto
against four baselines that vary only in the \emph{client partitioning scheme} and/or
the \emph{initialisation strategy}. Unless stated otherwise, the downstream federated
LSTM pipeline is kept identical across methods; only the grouping (client assignment)
differs.

\paragraph{Geo-based grouping (Geo-3 and Geo-7).}
Geo baselines cluster turbines purely in the planar coordinate space
$(\text{UTM}_x, \text{UTM}_y)$ and treat each geographic region as one federated client.
\textbf{Geo-3} selects $K=3$ by maximising the global silhouette score over candidate
$K$, yielding three spatial clients. \textbf{Geo-7} fixes $K=7$ to match the leaf count
of DRS-auto, enabling a fair comparison under the same number of groups while removing
behaviour features and recursive splitting.

\paragraph{Non-autoSplit Fed-KMeans (flat, feature-space, $K=6$).}
This baseline performs federated KMeans directly in the turbine behaviour feature space,
but without any recursive splitting: turbines are partitioned into $K$ clusters in a
single shot, and the resulting clusters are treated as federated clients for LSTM
training. To ensure comparability, we use the same federated hyperparameters as DRS-auto
(e.g., number of clients $n$ and communication rounds $c$), while selecting $K=6$ based on
an offline analysis balancing silhouette and SSE--$K$ behaviour.

\paragraph{K-means++ initialisation with the same recursive framework (K++-auto).}
To separate the impact of the initialisation strategy from the recursive framework, we
construct a baseline that replaces DRS initialisation with k-means++ while keeping the
entire recursive federated KMeans and auto-split procedure unchanged. This comparison
quantifies the additional benefit of DRS over k-means++ within the same hierarchical
pipeline.

\paragraph{Centralised training baseline.}
As a ``no partition, no federation'' framework, we train a single centralised LSTM by
merging all turbines within a representative behaviour cluster and evaluating it on the
same turbines. Due to compute constraints, we report this baseline on one representative
cluster, providing a reference point for the impact of federation on prediction accuracy and scalability.

\begin{figure}[!t]
\centering

\begin{subfigure}[b]{0.48\textwidth}
    \centering
    \includegraphics[height=10cm]{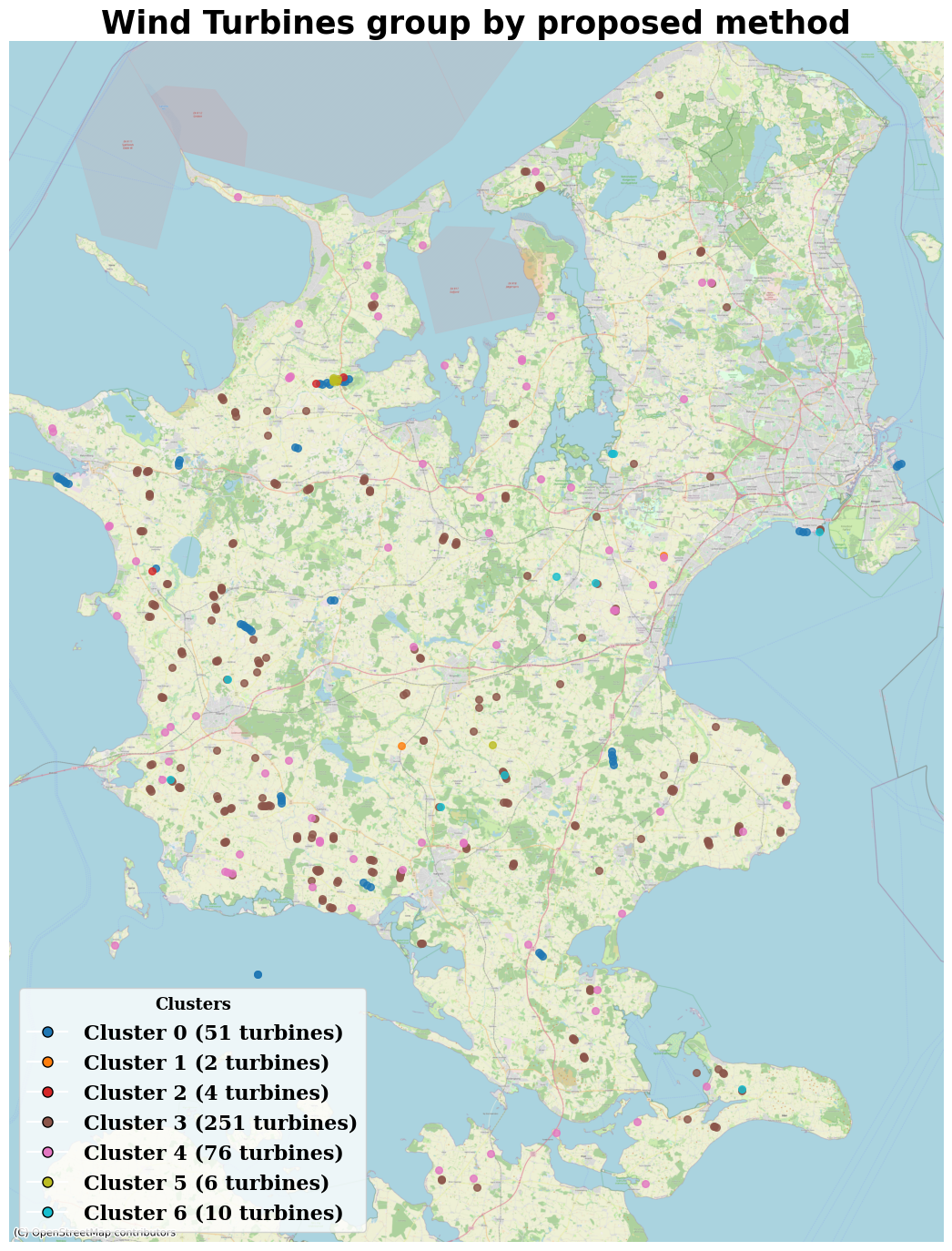}
    \caption{DRS-auto}
\end{subfigure}
\hspace{\fill}
\begin{subfigure}[b]{0.48\textwidth}
    \centering
    \includegraphics[height=10cm]{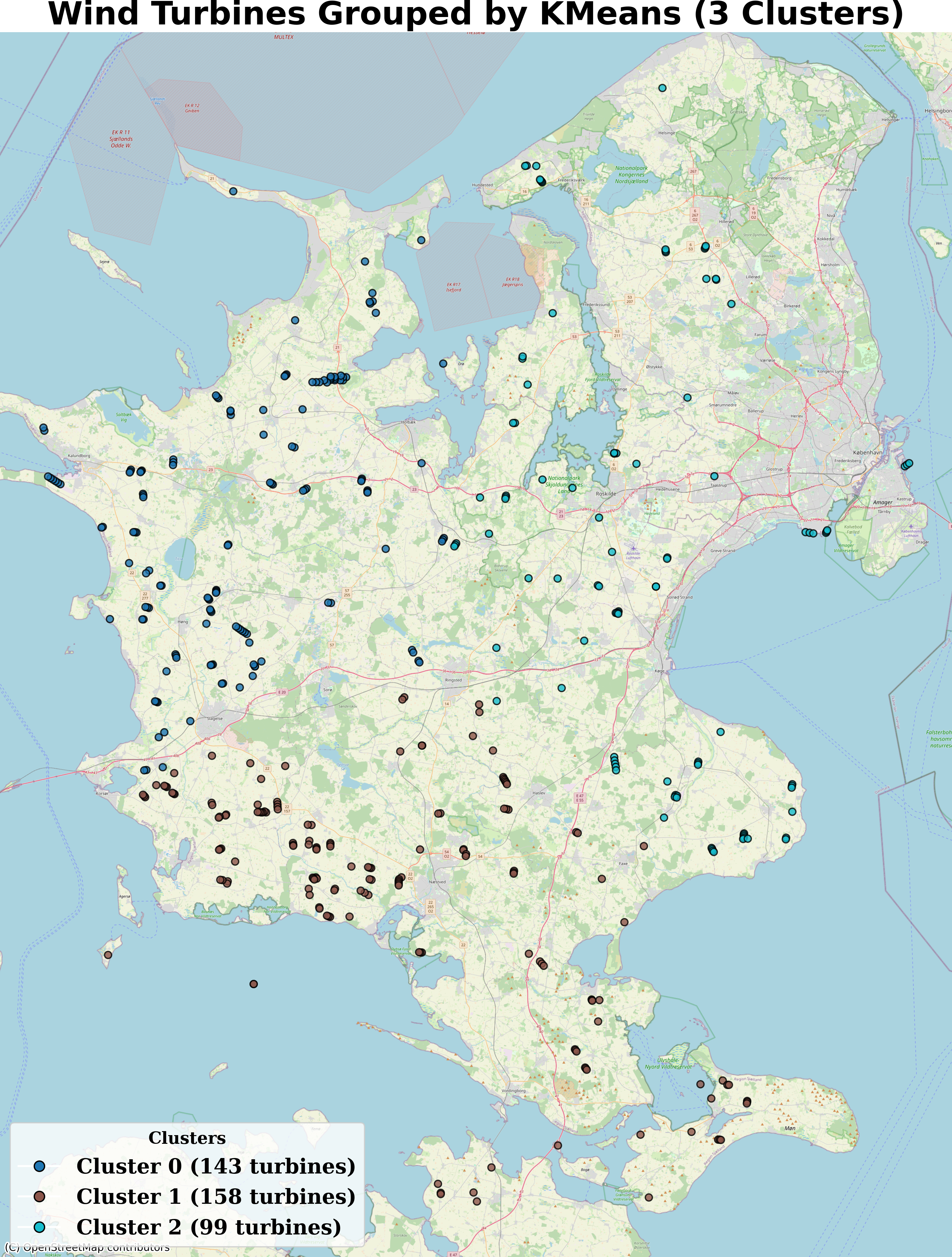}
    \caption{Geo-3}
\end{subfigure}

\vspace{0.6em}

\begin{subfigure}[b]{0.48\textwidth}
    \centering
    \includegraphics[height=10cm]{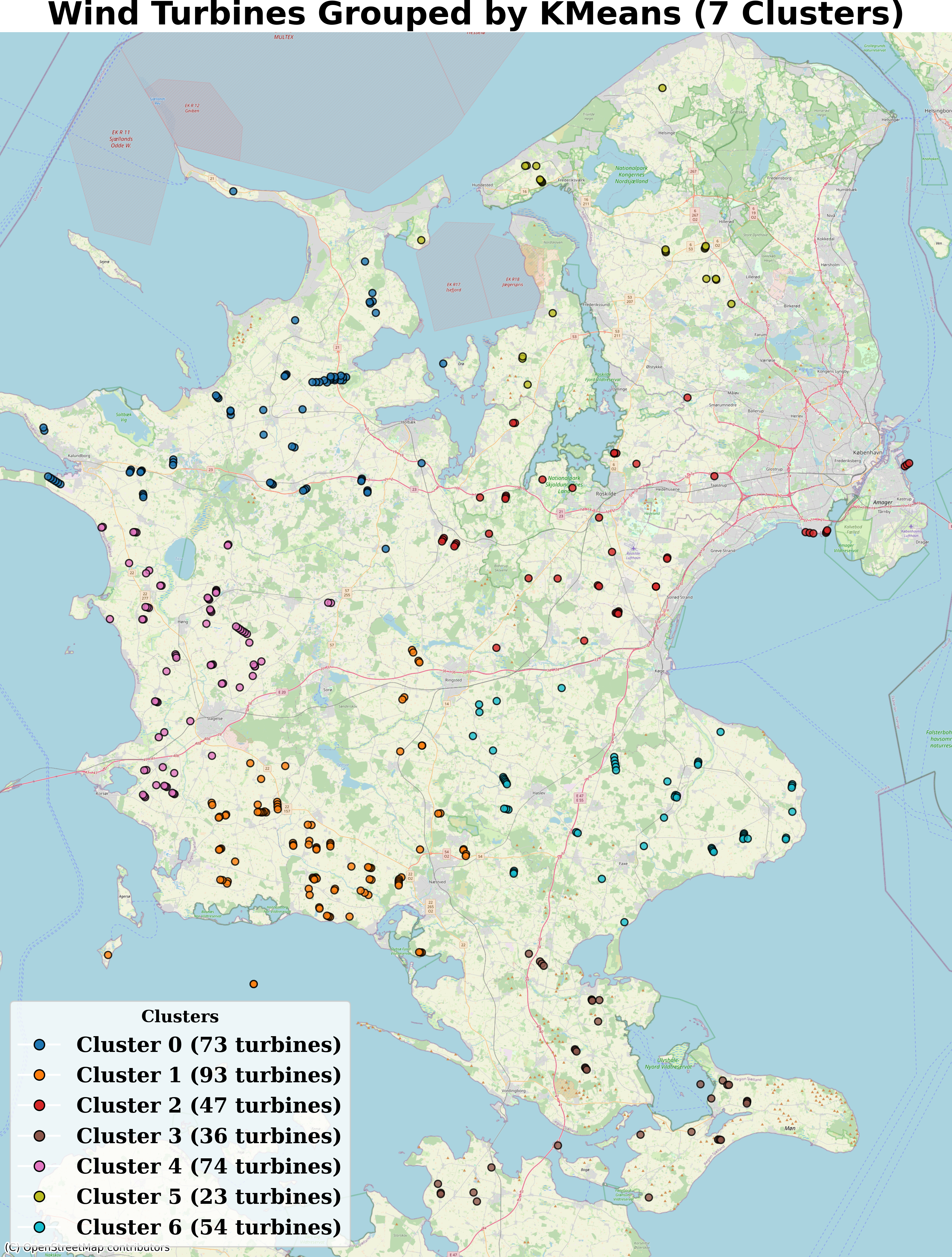}
    \caption{Geo-7}
\end{subfigure}
\hspace{\fill}
\begin{subfigure}[b]{0.48\textwidth}
    \centering
    \includegraphics[height=10cm]{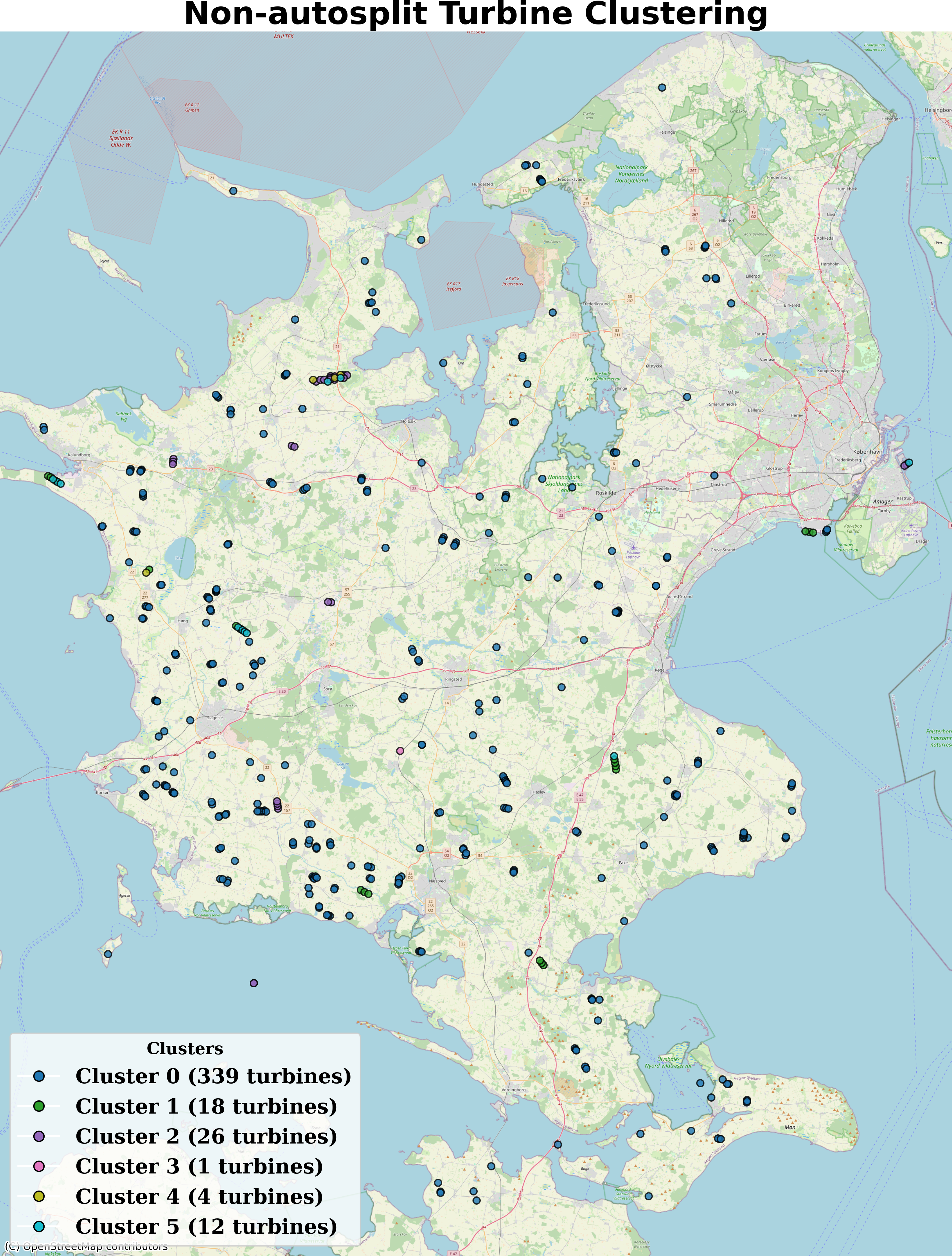}
    \caption{Non-autosplit}
\end{subfigure}

\caption{Comparison of behaviour- and geo-based grouping strategies.}
\label{fig:grouping_strategies_map}
\end{figure}

\textbf{Figure~\ref{fig:grouping_strategies_map}} visualizes the different client
grouping strategies considered on the same set of 400 turbines. In all panels, colours indicate cluster
membership.


\section{Results and Analysis}

\subsection{Clustering Results and Interpretation}

\subsubsection{Turbine behavior summary statistics}

\textbf{Table~\ref{tab:cluster-means}} reports, for each cluster, the number of turbines and
the mean value of each standardized feature. \textbf{Table~\ref{tab:cluster-stds}} shows the
corresponding within-cluster standard deviations.

\begin{table}[htbp]
    \centering
    \caption{Cluster-wise turbine counts and mean standardized feature values.}
    \label{tab:cluster-means}
    \setlength{\tabcolsep}{3pt} 
    \scriptsize                 
    \resizebox{\textwidth}{!}{%
    \begin{tabular}{r r r r r r r r}
        \toprule
        Cluster & Count &
        mean\_power\_scaled\_mean &
        std\_power\_scaled\_mean &
        cv\_scaled\_mean &
        zero\_ratio\_mean &
        ramp\_mean\_scaled\_mean &
        ramp\_std\_scaled\_mean \\
        \midrule
        0 &  51 &  2.3383 &  2.3408 & -0.1774 & 0.0672 & -1.1569 &  2.3493 \\
        1 &   2 & -0.7299 & -0.8549 & 13.2276 & 0.9987 &  0.2984 & -0.7925 \\
        2 &   4 &  1.7114 &  1.8085 & -0.1317 & 0.1102 &  4.7725 &  1.6934 \\
        3 & 251 & -0.3049 & -0.2922 & -0.0830 & 0.1465 &  0.0177 & -0.3018 \\
        4 &  76 & -0.6317 & -0.7048 &  0.0607 & 0.3157 &  0.2485 & -0.6684 \\
        5 &   6 &  0.5266 &  0.8178 & -0.0903 & 0.0900 &  0.7233 &  0.7399 \\
        6 &  10 & -0.3271 & -0.2911 & -0.0114 & 0.2397 &  1.1653 & -0.2892 \\
        \bottomrule
    \end{tabular}%
    }
\end{table}

\begin{table}[htbp]
    \centering
    \caption{Within-cluster standard deviations of standardized features.}
    \label{tab:cluster-stds}
    \setlength{\tabcolsep}{3pt}
    \scriptsize
    \resizebox{\textwidth}{!}{%
    \begin{tabular}{r r r r r r r}
        \toprule
        Cluster &
        mean\_power\_scaled\_std &
        std\_power\_scaled\_std &
        cv\_scaled\_std &
        zero\_ratio\_std &
        ramp\_mean\_scaled\_std &
        ramp\_std\_scaled\_std \\
        \midrule
        0 & 0.9186 & 0.8308 & 0.0209 & 0.0207 & 1.9099 & 0.8828 \\
        1 & 0.0000 & 0.0018 & 6.5609 & 0.0011 & 0.0000 & 0.0083 \\
        2 & 1.5020 & 1.0891 & 0.0810 & 0.0611 & 1.2589 & 0.9328 \\
        3 & 0.0971 & 0.1045 & 0.0484 & 0.0820 & 0.3357 & 0.0973 \\
        4 & 0.0736 & 0.1091 & 0.1783 & 0.1942 & 0.1157 & 0.0919 \\
        5 & 0.0610 & 0.1199 & 0.0168 & 0.0183 & 0.5113 & 0.1440 \\
        6 & 0.1615 & 0.1489 & 0.1734 & 0.2085 & 0.3848 & 0.1384 \\
        \bottomrule
    \end{tabular}%
    }
\end{table}

\subsubsection{Cluster-wise behavioural types}
\label{subsubsec:cluster_behaviour_types}

Based on the standardised feature statistics, the seven clusters represent distinct turbine behaviour types (\textbf{Table~\ref{tab:cluster_summary}}), which are detailed below:

\begin{table}[ht]
\centering
\caption{Summary of cluster behavioural types from DRS-auto.}
\label{tab:cluster_summary}
\setlength{\tabcolsep}{3pt}
\scriptsize
\begin{tabular}{clrl}
\toprule
\textbf{ID} & \textbf{Type} & \textbf{Size} & \textbf{Key Characteristics} \\
\midrule
0 & High power, high variability & 51 & Above-avg power ($\approx$2.34), low shutdown, high ramps \\
1 & Faulty / shutdown & 2 & Near-zero output, $\approx$100\% shutdown ratio (excluded) \\
2 & Ramp-dominated & 4 & Extreme ramping ($\approx$4.77), high heterogeneity \\
3 & Baseline stable & 251 & Near-average features, high homogeneity (main group) \\
4 & Mid-risk, low-output & 76 & Below-avg power, elevated shutdown ($\approx$0.32) \\
5 & Promising but volatile & 6 & Above-avg power, low shutdown, high variability \\
6 & Mildly unstable & 10 & Below-avg power, moderate ramping trend \\
\bottomrule
\end{tabular}
\end{table}

\begin{itemize}
    \item Cluster~0 turbines operate under favourable but ``rough'' wind conditions with high yield but challenging predictability. 
    \item Cluster~1 (two near-offline turbines) is excluded from forecasting as an anomaly group. 
    \item Cluster~2 exhibits extreme ramping behaviour useful for stress-testing models.
    \item Cluster~3, the dominant group, serves as the baseline with average behaviour and high internal homogeneity. 
    \item Cluster~4 turbines have low but smooth output, likely due to poor wind resource or curtailment. 
    \item Cluster~5 combines high energy contribution with pronounced dynamics. Cluster~6 shows moderate instability between stable and high-risk regimes.
\end{itemize}

In federated forecasting, Cluster~1 is excluded; the remaining six clusters (0, 2--6) serve as distinct client types for systematic performance evaluation.

\begin{figure}[htbp]
    \centering
    \begin{subfigure}[b]{0.48\textwidth}
        \centering
        \includegraphics[width=\linewidth]{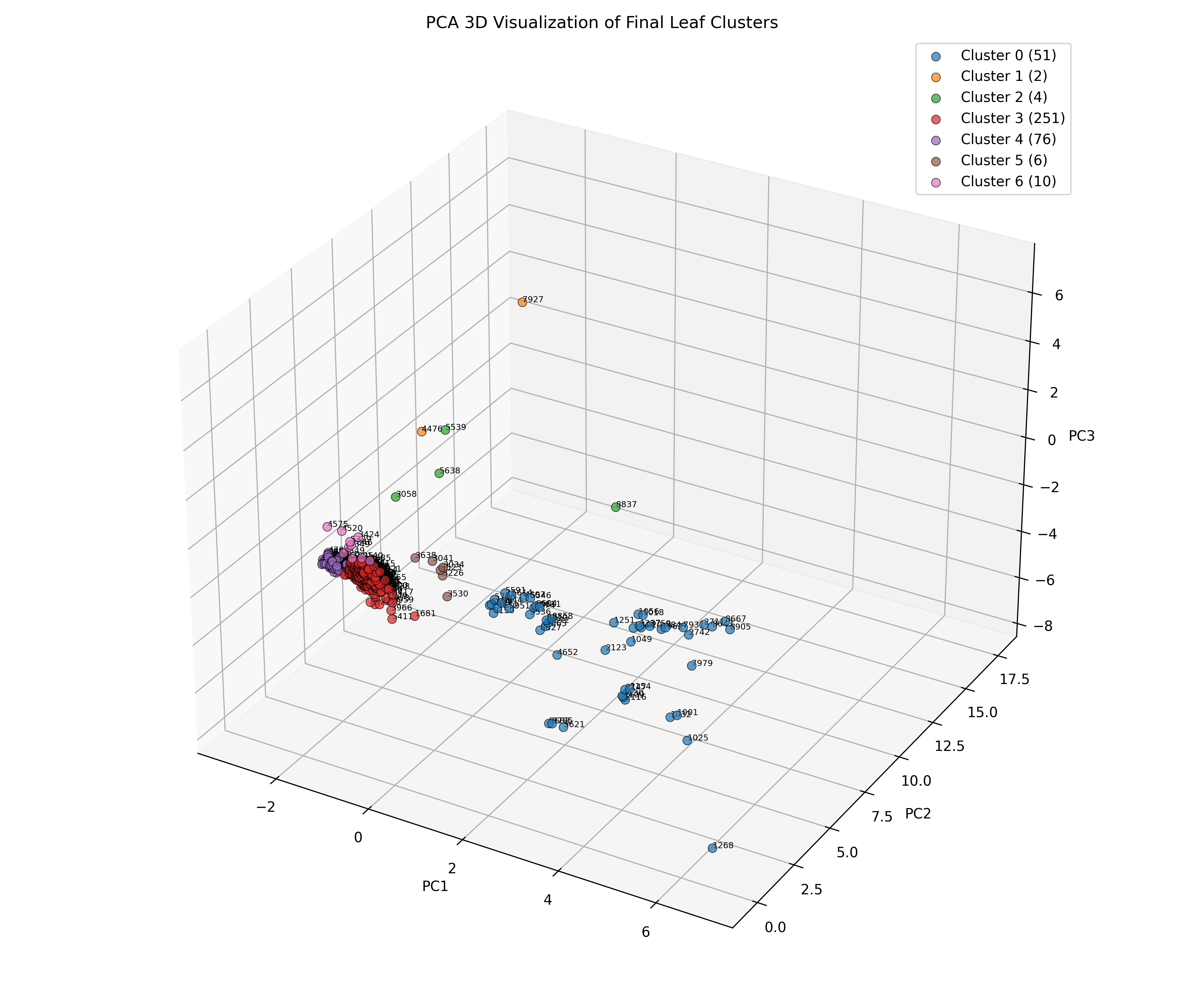}
        \caption{DRS-auto clustering in PCA}
        \label{fig:drs_pca}
    \end{subfigure}%
    \hfill
    \begin{subfigure}[b]{0.48\textwidth}
        \centering
        \includegraphics[width=\linewidth]{clustering/drs-map.png}
        \caption{DRS-auto clusters on the map}
        \label{fig:drs-auto_map}
    \end{subfigure}
    \caption{DRS-auto clustering results.}
    \label{fig:drs_results}
\end{figure}

Principal Component Analysis (PCA) projects the clustering features into 3D for visualisation~\cite{jolliffe2002principal}. \textbf{Figure~\ref{fig:drs_results}} shows DRS-auto clustering in feature space (panel~a) and geographical space (panel~b), confirming that behavioural clusters are not purely geographical but reflect local weather, siting, and operational strategies.

\subsection{Overall Prediction Performance Comparison Across Different Baselines}

\textbf{Tables~\ref{tab:baseline_train}} and~\textbf{\ref{tab:baseline_test}} summarize the
train- and test-set performance of all federated grouping strategies on the full
set of 400 turbines, while \textbf{Table~\ref{tab:drs_vs_centralized_51}} contrasts
DRS-auto with a centralized LSTM on the same 51-turbine subset. 

\begin{table}[ht]
  \centering\scriptsize
  \begin{minipage}[t]{0.48\textwidth}
    \centering
    \caption{Train-set performance of grouping strategies.}
    \label{tab:baseline_train}
    \begin{tabular}{lccccc}
      \hline
      Method & \#C & MSE & RMSE & MAE & $R^2$ \\
      \hline
      DRS-auto      & 7 & 0.0157 & \textbf{0.122} & 0.084 & 0.688 \\
      KMeans++-auto & 6 & \textbf{0.0157} & 0.122 & \textbf{0.084} & \textbf{0.689} \\
      Non-auto FK   & 6 & 0.0157 & 0.123 & 0.084 & 0.686 \\
      Geo-3         & 3 & 0.0160 & 0.123 & 0.085 & 0.681 \\
      Geo-7         & 7 & 0.0158 & 0.123 & 0.085 & 0.685 \\
      \hline
    \end{tabular}
  \end{minipage}
  \hfill
  \begin{minipage}[t]{0.48\textwidth}
    \centering
    \caption{Test-set performance of grouping strategies.}
    \label{tab:baseline_test}
    \begin{tabular}{lccccc}
      \hline
      Method & \#C & MSE & RMSE & MAE & $R^2$ \\
      \hline
      DRS-auto      & 7 & \textbf{0.0179} & \textbf{0.130} & \textbf{0.088} & 0.699 \\
      KMeans++-auto & 6 & 0.0180 & 0.130 & 0.088 & 0.700 \\
      Non-auto FK   & 6 & 0.0180 & 0.131 & 0.089 & \textbf{0.715} \\
      Geo-3         & 3 & 0.0655 & 0.142 & 0.091 & 0.421 \\
      Geo-7         & 7 & 0.0654 & 0.142 & 0.091 & 0.467 \\
      \hline
    \end{tabular}
  \end{minipage}
\end{table}

\begin{table}[ht]
  \centering\scriptsize
  \caption{Comparison between federated DRS-auto and centralized LSTM on the same 51-turbine cluster.\protect\footnotemark}
  \label{tab:drs_vs_centralized_51}
  \begin{tabular}{lccccc}
    \hline
    Method                 & Split  & MSE      & RMSE     & MAE      & $R^2$    \\
    \hline
    DRS-auto (federated)   & Train  & 0.028091 & 0.166485 & 0.122107 & 0.678116 \\
    DRS-auto (federated)   & Test   & 0.028822 & 0.168435 & 0.121883 & 0.711436 \\
    Centralized LSTM       & Train  & 0.006223 & 0.078890 & 0.050925 & 0.887562 \\
    Centralized LSTM       & Test   & 0.006159 & 0.078483 & 0.049881 & 0.905096 \\
    \hline
  \end{tabular}
\end{table}
\footnotetext{Due to training-time and resource constraints, the centralized baseline was not run on all 400 turbines; instead, both models are trained and evaluated on the same representative subset of 51 turbines. DRS-auto keeps the federated constraint (multiple logical clients), whereas the centralized LSTM trains a single global model on all available data within this subset.}

From these results, the relative performance tiers in the fully federated setting
(all 400 turbines) can be summarised as follows.

\paragraph{DRS-auto and KMeans++-auto: top-tier automatic grouping.}
On the full 400-turbine setting, DRS-auto and KMeans++-auto achieve essentially the
same predictive performance (\textbf{Table~\ref{tab:baseline_train}}) and (\textbf{Table~\ref{tab:baseline_test}}). The key difference is not accuracy
but structure: KMeans++-auto yields a one-shot partition obtained from a centralised
feature-matrix search, whereas DRS-auto performs recursive auto-splitting with explicit
stopping rules, producing a balanced set of seven leaf clusters organised as a
hierarchical tree. This hierarchy is operationally useful (e.g., for outlier-cluster
removal and cluster-specific adaptation) while preserving privacy, since clients only
share aggregated centres and counts.

\paragraph{Non-autoSplit Fed-KMeans: competitive averages but limited structure.}
Non-autoSplit Fed-KMeans can appear close in aggregate metrics, but its one-shot flat
partition lacks hierarchical refinement and typically results in less balanced and less
homogeneous groups. This reduces interpretability and limits downstream flexibility,
highlighting that in federated forecasting the \emph{structure} of grouping matters, not
only the overall error level.

\paragraph{Geo-based grouping (Geo-3/Geo-7): intuitive but weaker under heterogeneity.}
Geo baselines group turbines solely by spatial proximity in $(\mathrm{UTM}_x,\mathrm{UTM}_y)$.
While training errors may remain comparable, test performance is consistently lower,
especially for Geo-3, indicating that geographic proximity alone mixes behaviourally
different turbines within the same client. Geo-7 improves slightly by increasing the
number of groups, but still lags behind behaviour-based grouping.

\paragraph{Centralised training (51-turbine subset): an optimistic upper bound.}
On a representative subset of 51 turbines (\textbf{Table~\ref{tab:drs_vs_centralized_51}}), a
centralised LSTM substantially outperforms DRS-auto. This comparison should be treated
as an optimistic upper-bound reference: it ignores data locality and privacy constraints
and is not directly comparable to the full 400-turbine federated setting.

\paragraph{Summary.}
Overall, the experiments indicate that (i) grouping strategy is more important than
simply matching the number of clients, (ii) DRS-auto matches the strongest automatic
baseline in accuracy while producing a more balanced and hierarchical partition that
supports practical extensions, and (iii) purely geographic partitioning is a useful
sanity baseline but is clearly suboptimal for highly heterogeneous turbine fleets.
\subsection{Cluster-wise Forecast Performance under DRS-auto Grouping}
\label{subsec:cluster_forecast_metrics}

\begin{figure}[H]
    \centering
    \begin{subfigure}[b]{0.45\textwidth}
        \centering
        \includegraphics[width=\linewidth]{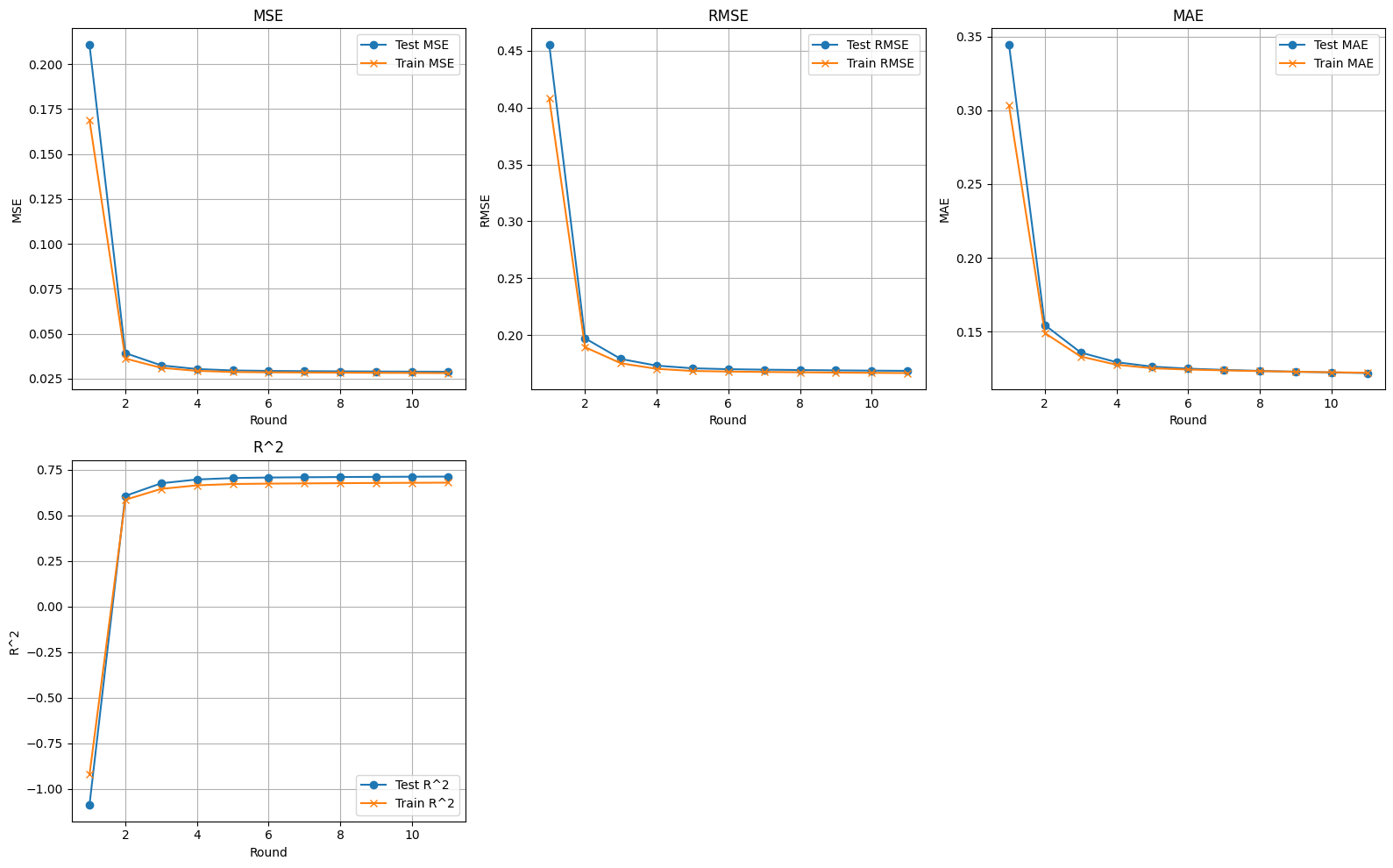}
        \caption{Cluster 0}
        \label{fig:metrics_cluster0}
    \end{subfigure}\hfill
    \begin{subfigure}[b]{0.45\textwidth}
        \centering
        \includegraphics[width=\linewidth]{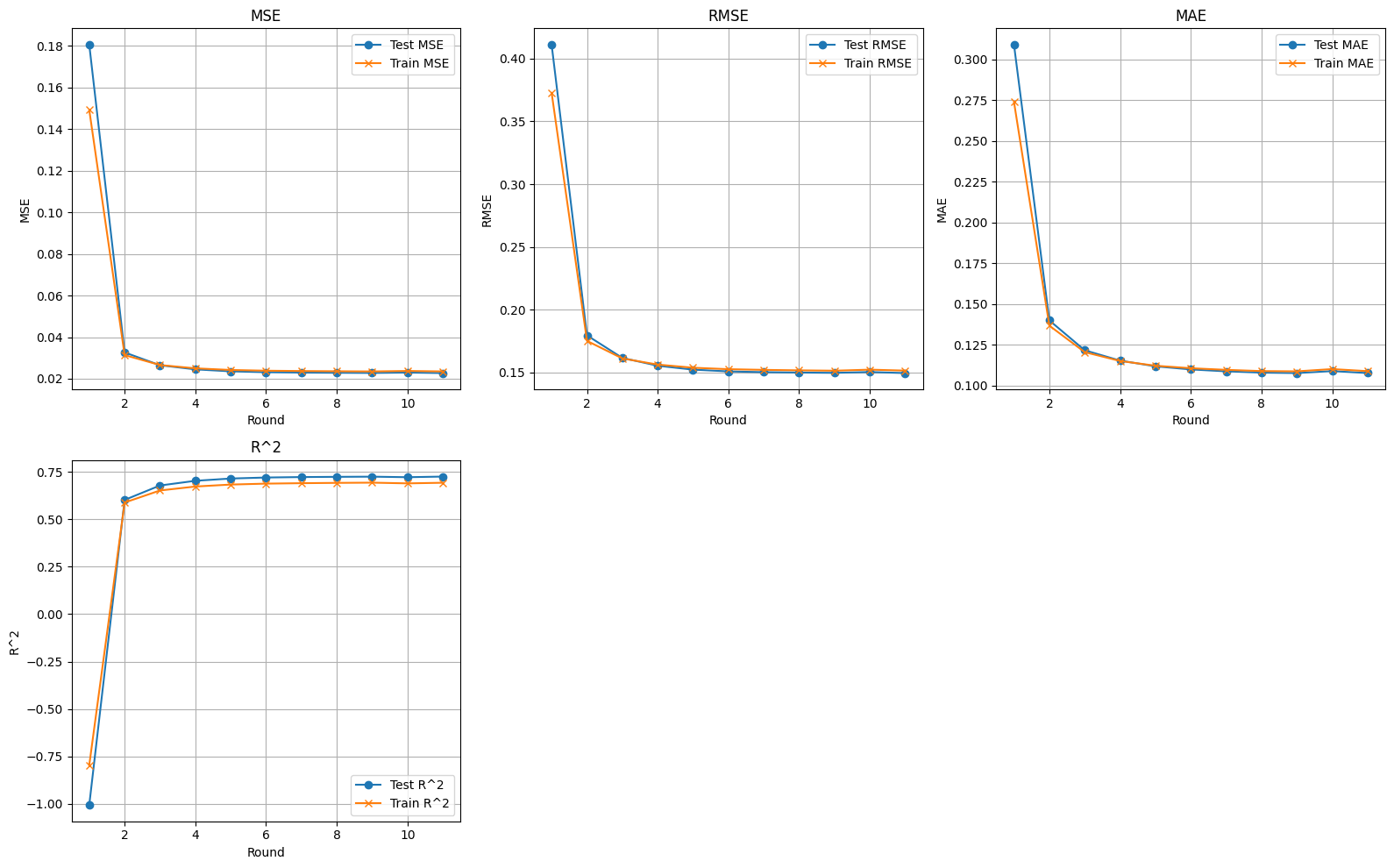}
        \caption{Cluster 2}
        \label{fig:metrics_cluster2}
    \end{subfigure}

    \vspace{0.5\baselineskip}

    \begin{subfigure}[b]{0.45\textwidth}
        \centering
        \includegraphics[width=\linewidth]{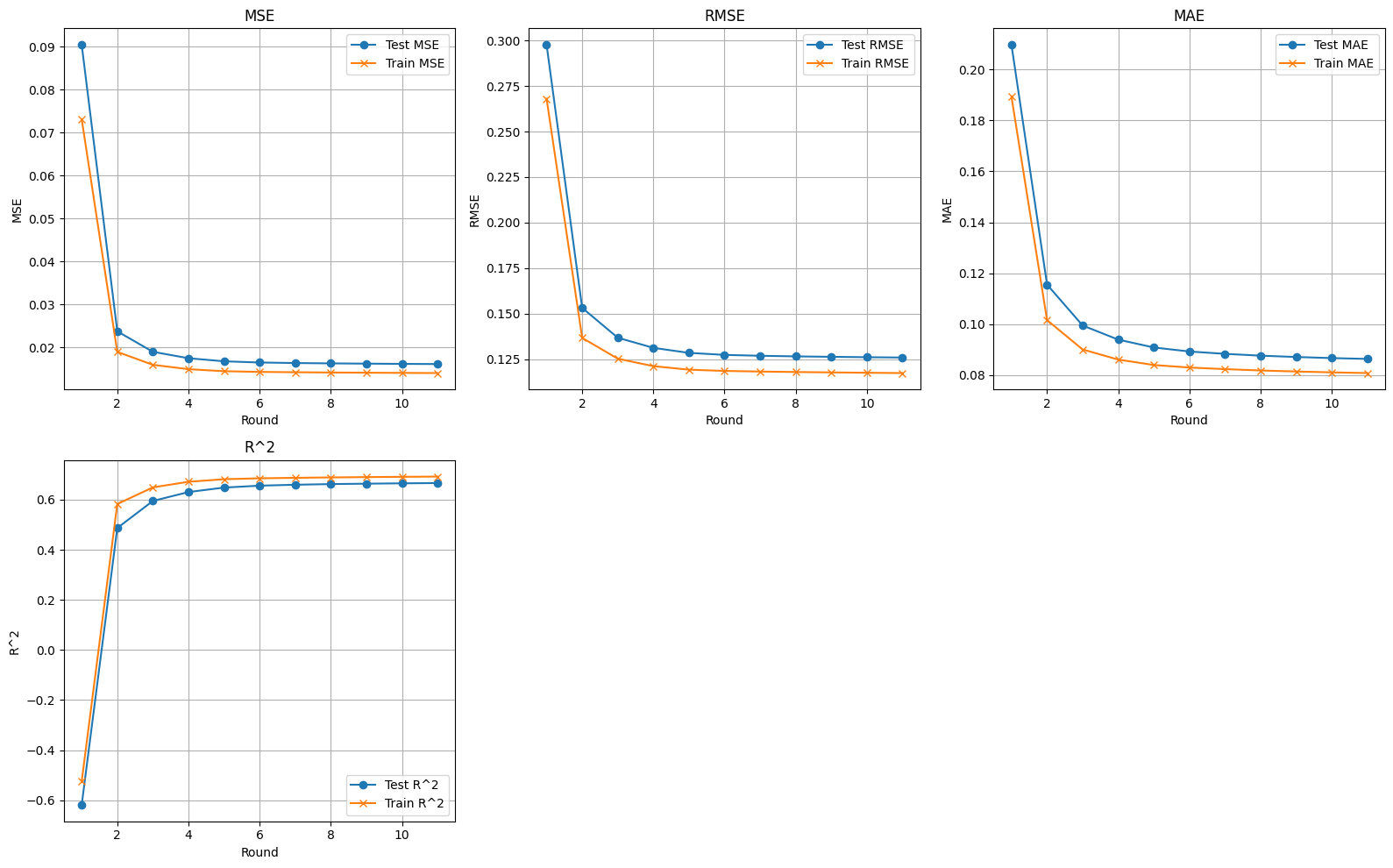}
        \caption{Cluster 3}
        \label{fig:metrics_cluster3}
    \end{subfigure}\hfill
    \begin{subfigure}[b]{0.45\textwidth}
        \centering
        \includegraphics[width=\linewidth]{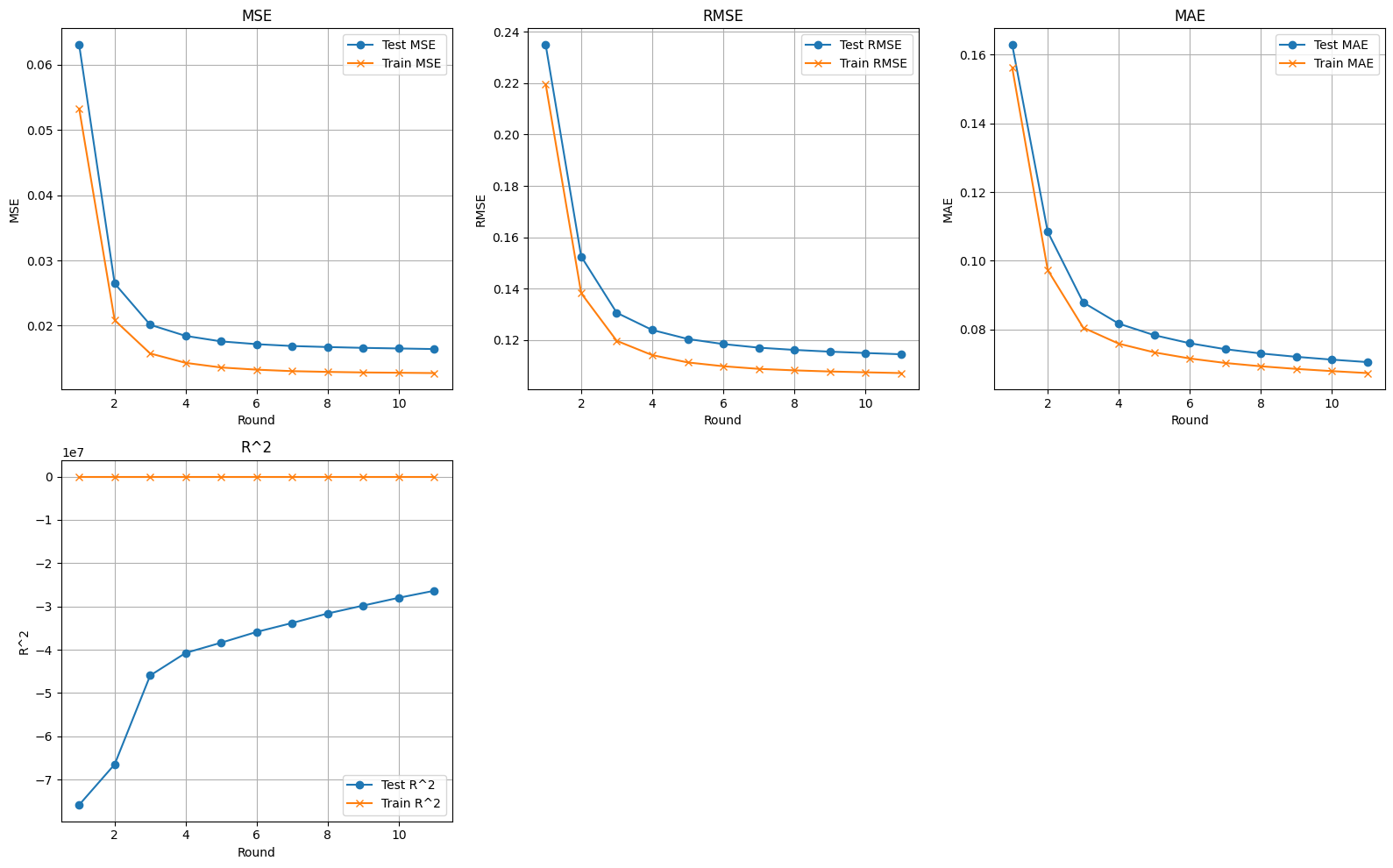}
        \caption{Cluster 4}
        \label{fig:metrics_cluster4}
    \end{subfigure}

    \vspace{0.5\baselineskip}

    \begin{subfigure}[b]{0.45\textwidth}
        \centering
        \includegraphics[width=\linewidth]{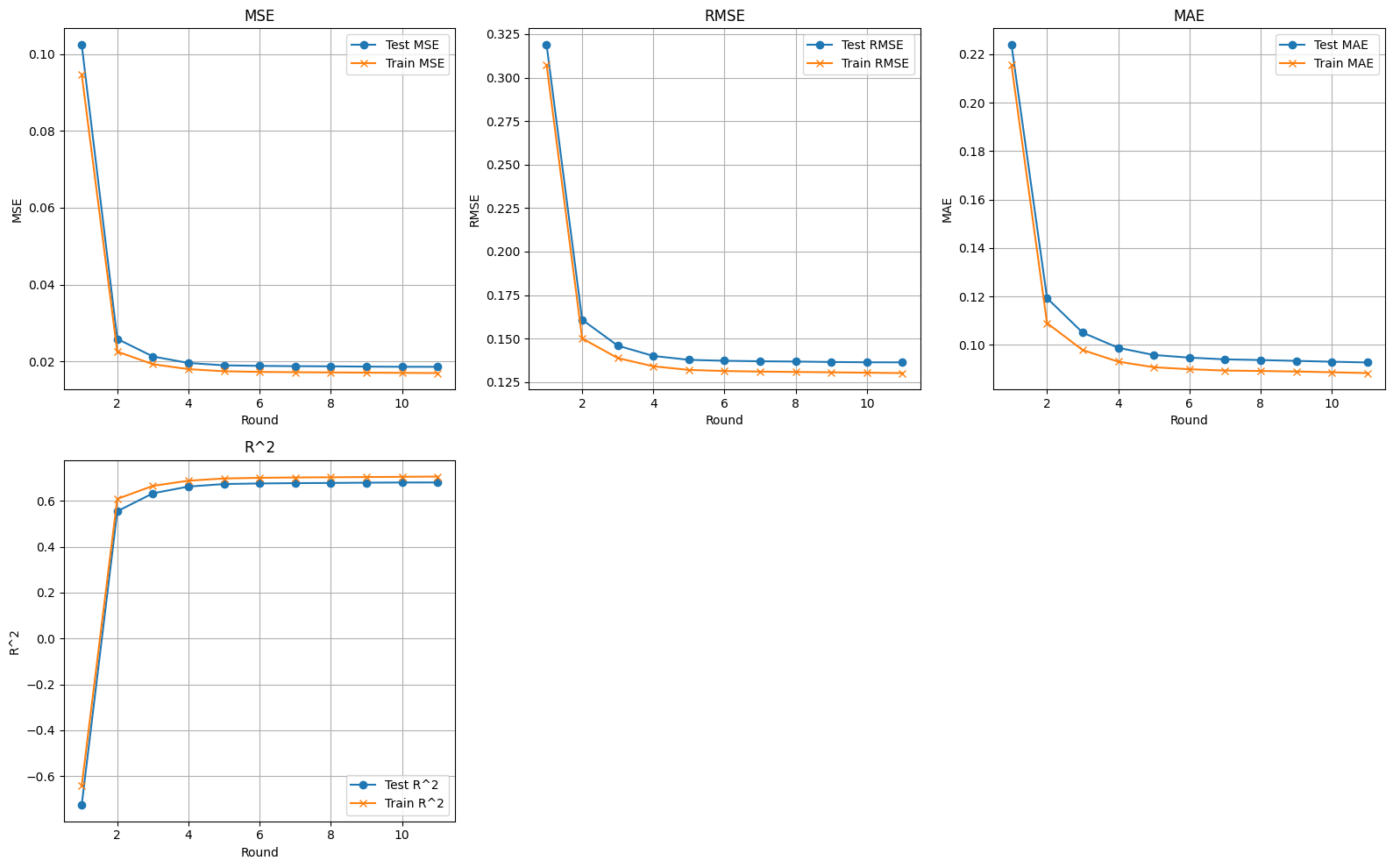}
        \caption{Cluster 5}
        \label{fig:metrics_cluster5}
    \end{subfigure}\hfill
    \begin{subfigure}[b]{0.45\textwidth}
        \centering
        \includegraphics[width=\linewidth]{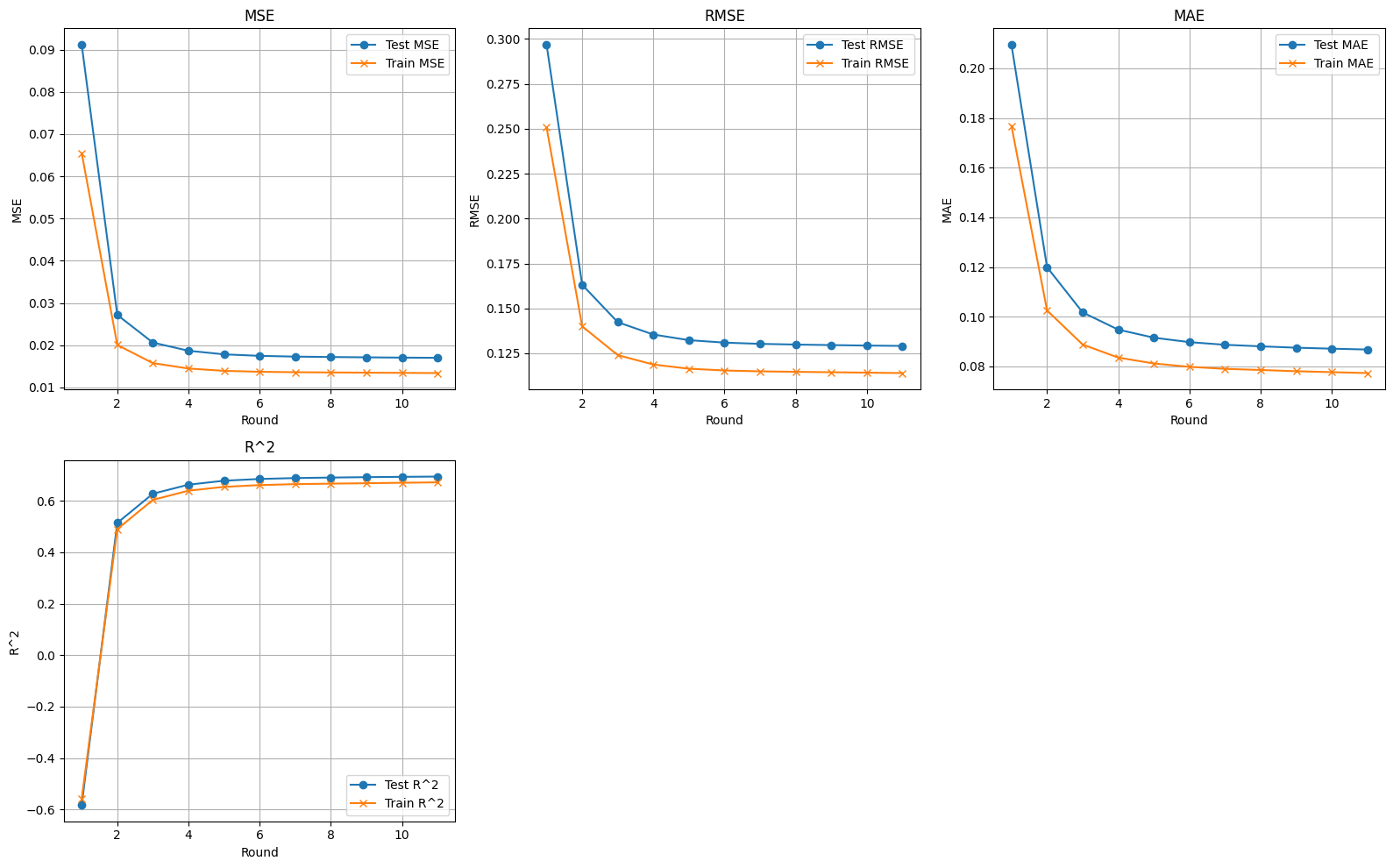}
        \caption{Cluster 6}
        \label{fig:metrics_cluster6}
    \end{subfigure}

    \caption{Forecast metrics per cluster under DRS-auto grouping.}
    \label{fig:cluster_metrics_forecasts}
\end{figure}

\textbf{Figure~\ref{fig:cluster_metrics_forecasts}} reports the evolution of the main regression metrics (MSE, RMSE, MAE,
and $R^2$) over federated training rounds for each behaviour cluster under the DRS-auto
grouping. Across all clusters, both train and test losses drop rapidly during the first few
communication rounds and then stabilise on a plateau, with training and test curves
closely aligned. This indicates that the federated LSTM does not suffer from severe
overfitting in any cluster; the differences lie mainly in the final error levels and in how
quickly and smoothly the $R^2$ curves saturate.

Tables~\ref{tab:cluster_train_metrics} and~\ref{tab:cluster_test_metrics} summarise the
final-round metrics per cluster. The baseline stable group (Cluster~3) attains the lowest
MSE and MAE, reflecting that turbines with regular, moderate behaviour are easiest to
predict. The ramp-dominated and mildly unstable groups (Clusters~2 and~6) reach
slightly higher error levels but still achieve $R^2 \approx 0.72$–$0.73$ on the test set,
showing that the model can capture most of their variability despite stronger dynamics.
In contrast, the mid-risk, low-output group (Cluster~4) exhibits the weakest $R^2$,
consistent with its large fraction of shutdown periods and low variance, which makes
relative-fit measures less stable. The high-power, high-variability and “promising but
volatile’’ groups (Clusters~0 and~5) fall in between: their absolute errors are modest,
but increased volatility leads to somewhat higher MSE and RMSE compared to the
baseline cluster.

\begin{table}[ht]
  \centering\scriptsize
  \begin{minipage}[t]{0.48\textwidth}
    \centering
    \caption{Train metrics under DRS-auto (last round).}
    \label{tab:cluster_train_metrics}
    \begin{tabular}{lcccc}
      \hline
      Cluster & MSE & RMSE & MAE & $R^2$ \\
      \hline
      0 & 0.02809 & 0.16649 & 0.12211 & 0.67812 \\
      2 & 0.02351 & 0.15157 & 0.10882 & 0.69281 \\
      3 & 0.01399 & 0.11749 & 0.08085 & 0.69233 \\
      4 & 0.01274 & 0.10720 & 0.06727 & 0.23888 \\
      5 & 0.01699 & 0.13016 & 0.08832 & 0.70551 \\
      6 & 0.01340 & 0.11401 & 0.07734 & 0.67153 \\
      \hline
    \end{tabular}
  \end{minipage}
  \hfill
  \begin{minipage}[t]{0.48\textwidth}
    \centering
    \caption{Test metrics under DRS-auto (last round).}
    \label{tab:cluster_test_metrics}
    \begin{tabular}{lcccc}
      \hline
      Cluster & MSE & RMSE & MAE & $R^2$ \\
      \hline
      0 & 0.02882 & 0.16843 & 0.12188 & 0.71144 \\
      2 & 0.02272 & 0.14961 & 0.10766 & 0.72539 \\
      3 & 0.01609 & 0.12601 & 0.08640 & 0.66642 \\
      4 & 0.01642 & 0.11453 & 0.07047 & $-2.64\times 10^{7}$ \\
      5 & 0.01862 & 0.13632 & 0.09270 & 0.67982 \\
      6 & 0.01701 & 0.12913 & 0.08682 & 0.69387 \\
      \hline
    \end{tabular}
  \end{minipage}

  \vspace{0.3em}
  {\raggedright\footnotesize
  Note: Cluster~1 is excluded from these tables because it consists of two turbines
  identified as almost permanently shut down (faulty anomaly cluster). Cluster~4
  contains a comparatively large share of turbines with irregular production. For
  practical reasons, no strict outlier filtering was applied at the modelling stage in
  order to keep as many turbines as possible in the federated setting; instead, this
  cluster is interpreted as a mid-risk group that can be targeted for customised
  treatment or separate models in future work (see Section~\ref{subsec:cluster4_finetune}).
  \par}
\end{table}

\subsection{24-Hour Forecast Case for Representative Turbines}

To obtain a more intuitive view of the short-term forecasting performance, this work also presents 24-hour forecast case studies for representative turbines. Specifically, for each cluster obtained by DRS-auto, one turbine is randomly selected, and its time series is used to plot the next 24 hours of true power and predicted power. By comparing representative turbines from different clusters and production levels, we can visually assess how well the federated LSTM model under DRS-auto grouping tracks day-ahead dynamics.

\begin{figure}[H]
    \centering
    \begin{subfigure}[b]{0.32\textwidth}
        \centering
        \includegraphics[width=\linewidth]{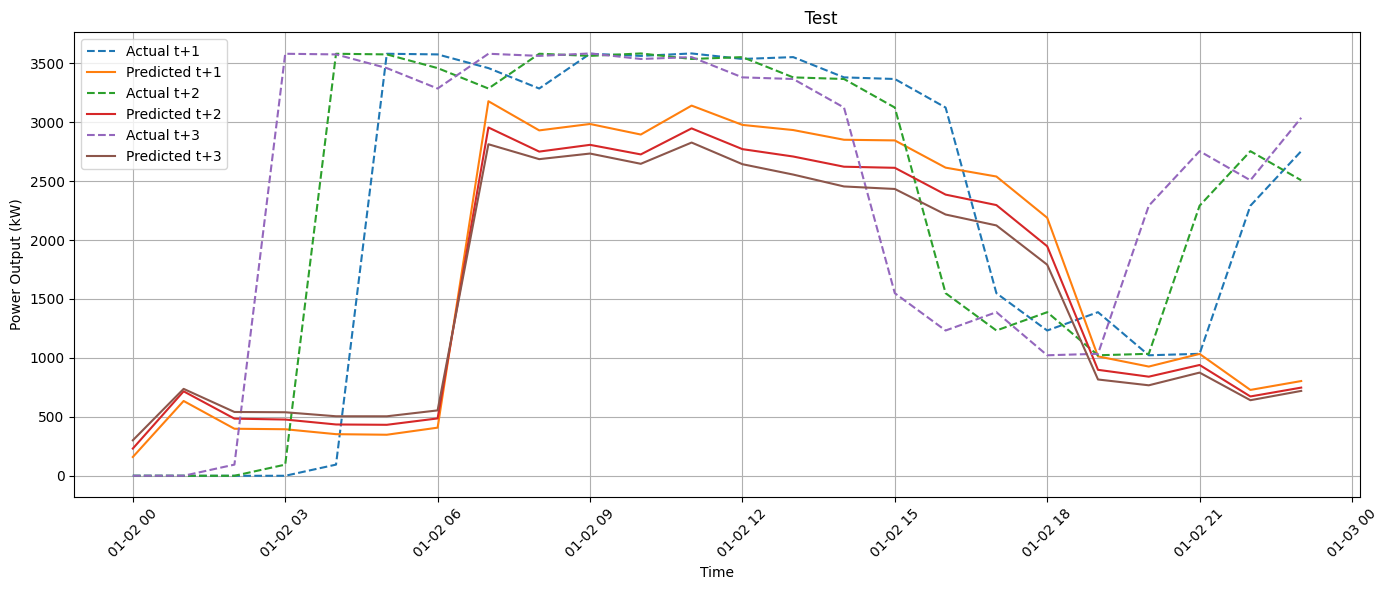}
        \caption{Cluster 0}
        \label{fig:forecast_cluster0}
    \end{subfigure}\hfill
    \begin{subfigure}[b]{0.32\textwidth}
        \centering
        \includegraphics[width=\linewidth]{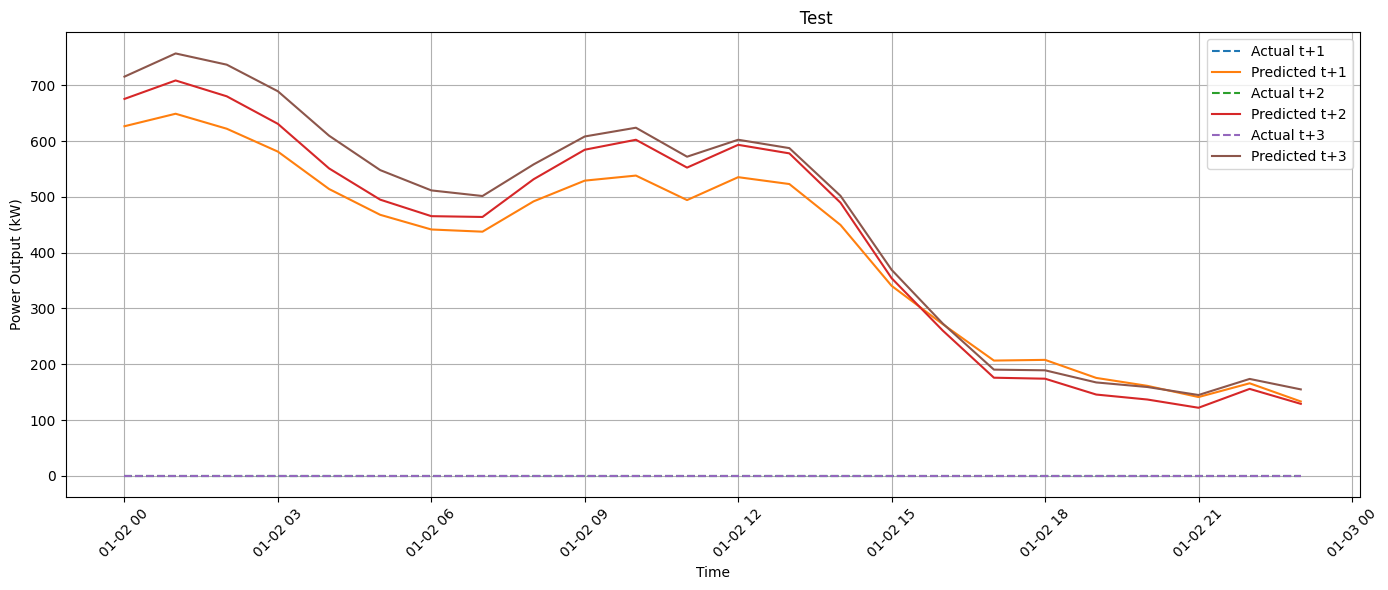}
        \caption{Cluster 2}
        \label{fig:forecast_cluster2}
    \end{subfigure}\hfill
    \begin{subfigure}[b]{0.32\textwidth}
        \centering
        \includegraphics[width=\linewidth]{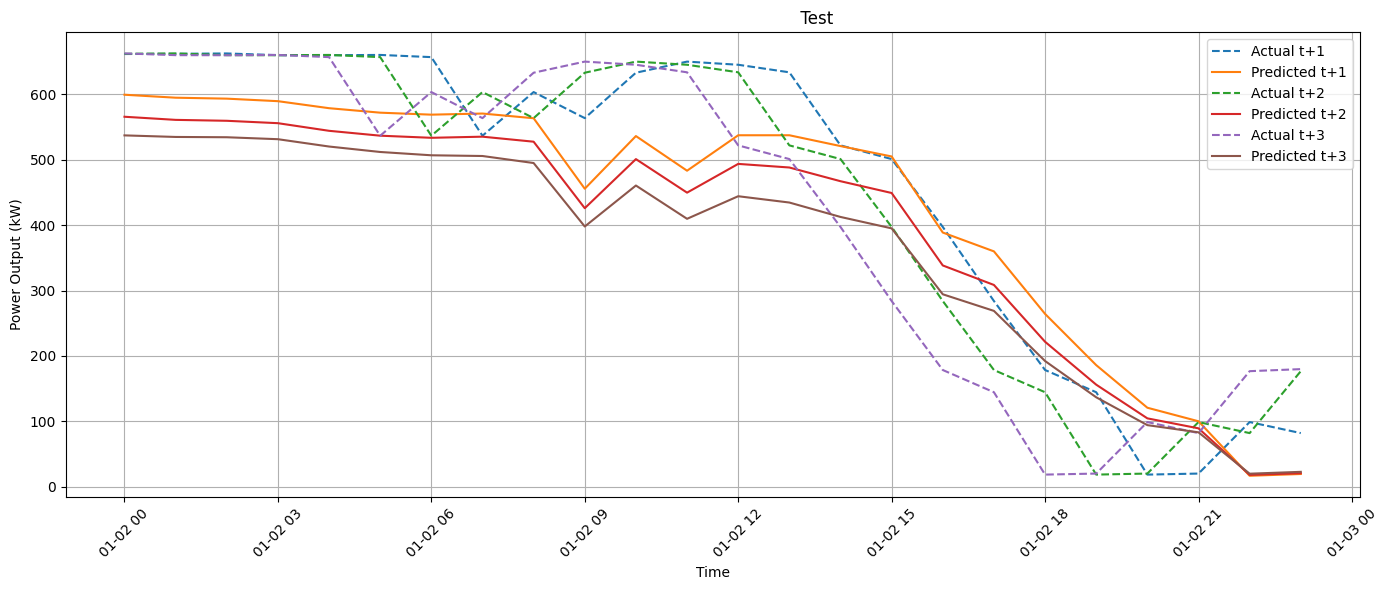}
        \caption{Cluster 3}
        \label{fig:forecast_cluster3}
    \end{subfigure}

    \vspace{0.5\baselineskip}

    \begin{subfigure}[b]{0.32\textwidth}
        \centering
        \includegraphics[width=\linewidth]{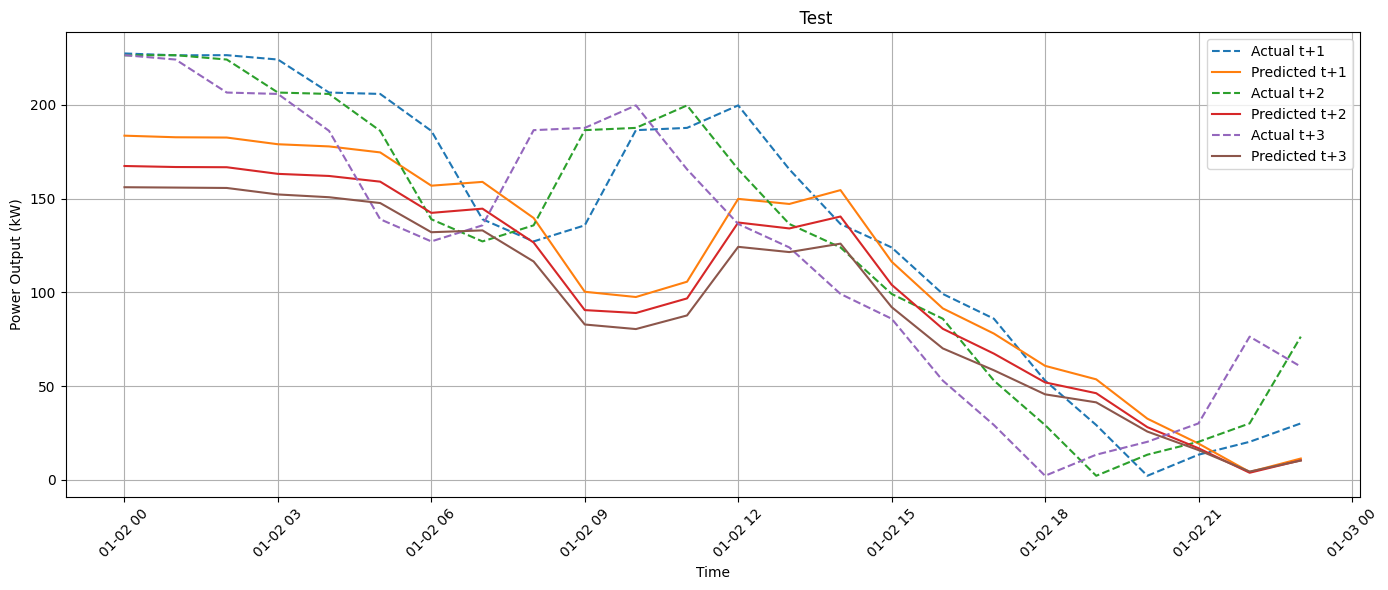}
        \caption{Cluster 4}
        \label{fig:forecast_cluster4}
    \end{subfigure}\hfill
    \begin{subfigure}[b]{0.32\textwidth}
        \centering
        \includegraphics[width=\linewidth]{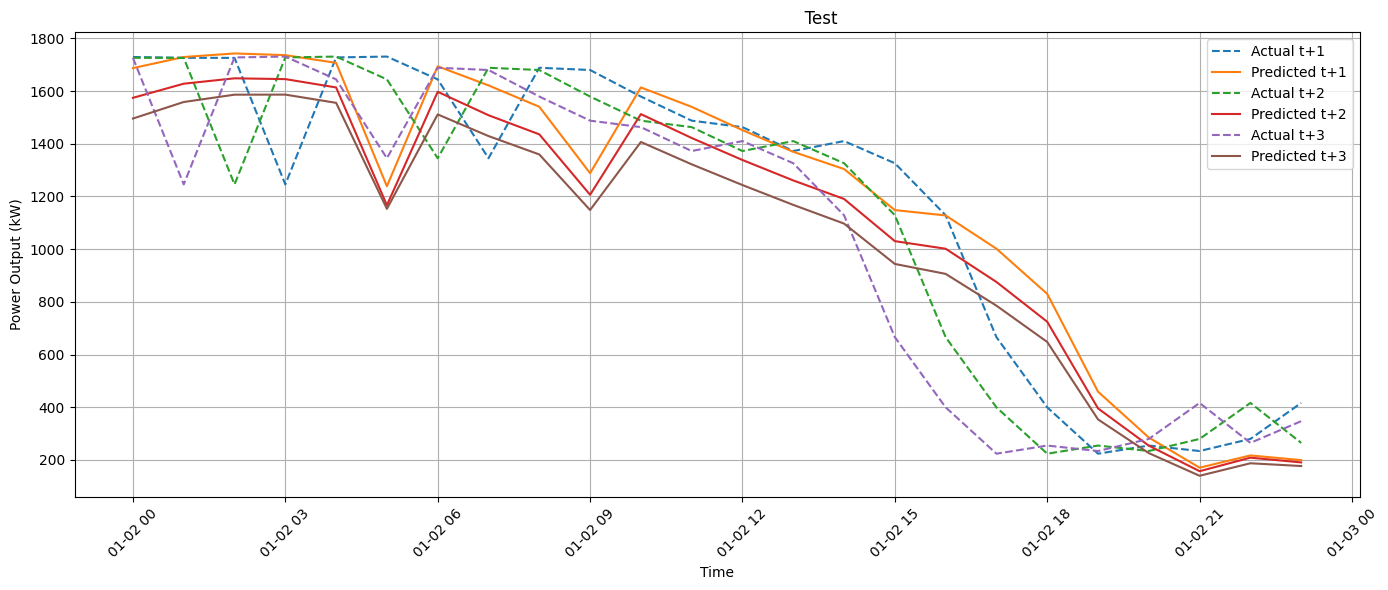}
        \caption{Cluster 5}
        \label{fig:forecast_cluster5}
    \end{subfigure}\hfill
    \begin{subfigure}[b]{0.32\textwidth}
        \centering
        \includegraphics[width=\linewidth]{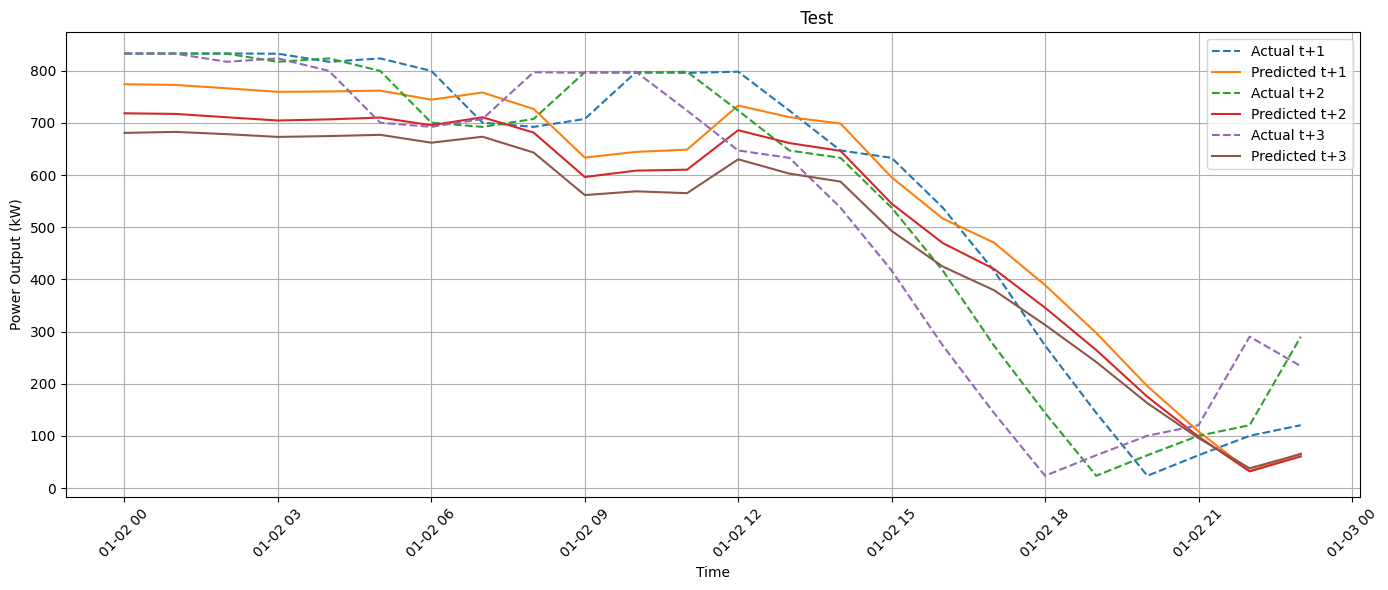}
        \caption{Cluster 6}
        \label{fig:forecast_cluster6}
    \end{subfigure}

    \caption{24-hour forecast cases for one representative turbine per cluster under DRS-auto grouping.}
    \label{fig:cluster_24h_forecasts}
\end{figure}

From the 24-hour forecasts in Fig.~\ref{fig:cluster_24h_forecasts}, where the \textbf{solid line} denotes the \emph{predicted} power and the \textbf{dashed line} denotes the \emph{true} power output, we observe that the model generally follows the main daily trend, but with clear differences across clusters. For Cluster~0 and Cluster~5 (high-output, high-variability groups), the model reproduces the morning/evening ramps and daytime high-production period reasonably well, but during late-night or early-morning episodes of sharp fluctuations, the predicted curve is visibly smoother than the dashed curve, with local spikes and sudden changes partially “flattened out”. Cluster~2 shows the overall best alignment between solid and dashed lines, indicating that, despite strong ramps and volatility, the model has captured the average dynamics of this behaviour group particularly well. Cluster~3, which contains the majority of turbines and represents the baseline stable group, exhibits good agreement between prediction and reality over morning ramps, mid-day plateau and evening decline, and can be regarded as the reference performance on “typical” turbines. 

In contrast, Cluster~4 is a mid-risk group with many shutdowns: these turbines may not only stop at night but also frequently face curtailment, maintenance or half-day shutdowns during daytime. Cluster~6 is mildly unstable, with occasional additional shutdowns or irregular variations. 

For Cluster~4 and Cluster~6, the owner’s short-term operational or curtailment decisions on a specific day (for example, whether to produce more or less around noon) can deviate from the cluster’s average behaviour pattern learned by the model. Since the plots are generated from a \emph{single randomly selected turbine} per cluster, any advance or delay in that turbine’s actual “daily schedule” relative to the cluster-level average can lead to situations around midday where the predicted and true curves move in opposite directions and cross each other, forming a small loop. This illustrates that, especially in shutdown- and curtailment-heavy clusters, there remains room for further improvement via cluster- or turbine-level customised fine-tuning.

\subsection{Cluster-Specific Fine-Tuning for the Mid-Risk Group (Cluster 4)}
\label{subsec:cluster4_finetune}

DRS-auto identifies Cluster~4 as a \emph{mid-risk, low-output} group. In practice, this
cluster also contains extreme cases (near-always-off or quasi-constant power series),
which yield near-zero target variance for some clients and lead to unstable $R^2$ and an
atypical error profile. To improve model stability without changing the clustering, we
perform a lightweight cluster-specific fine-tuning step by filtering out such
uninformative clients within Cluster~4.

Let $y_{\text{val}}$ denote the validation target series of a turbine client. We exclude a
client if any of the following holds:
\[
\forall t:\, y_{\text{val}}(t)=0
\quad \lor \quad
\max(y_{\text{val}})<0.1
\quad \lor \quad
\operatorname{std}(y_{\text{val}})<0.05.
\]
The federated LSTM is then fine-tuned on the remaining clients using the same model
architecture and hyperparameters.

\begin{figure}[t]
    \centering
    \begin{subfigure}[b]{0.49\textwidth}
        \centering
        \includegraphics[width=\linewidth]{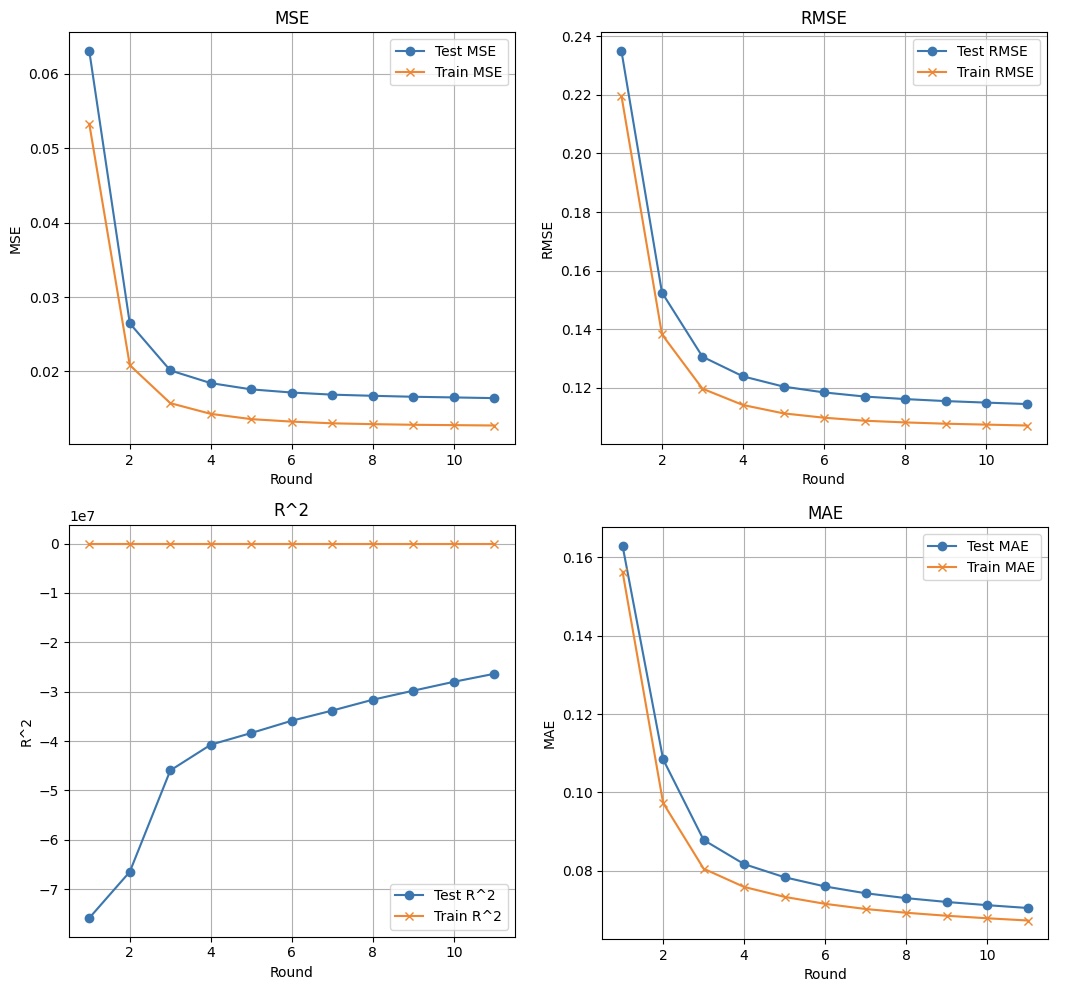}
        \caption{Before fine-tuning}
        \label{fig:cluster4_before}
    \end{subfigure}
    \hfill
    \begin{subfigure}[b]{0.49\textwidth}
        \centering
        \includegraphics[width=\linewidth]{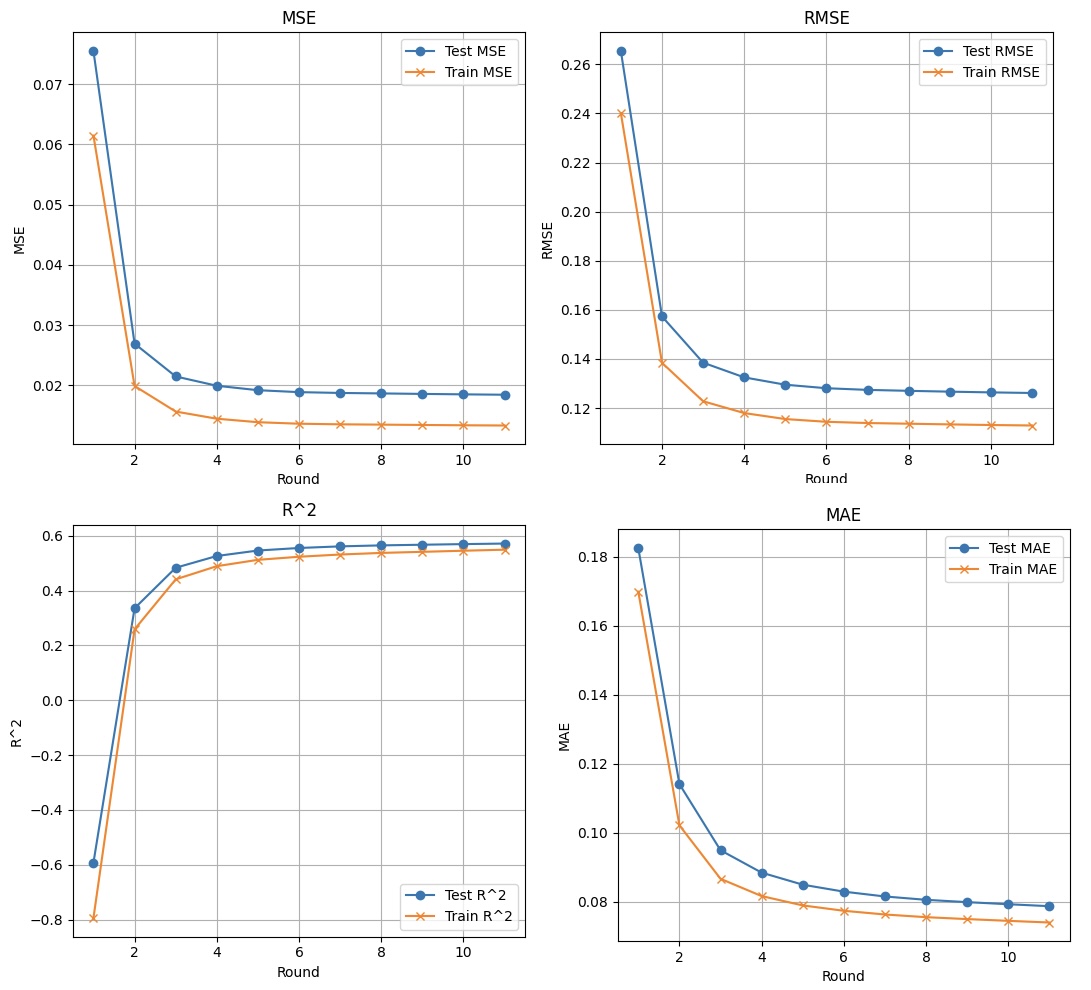}
        \caption{After fine-tuning}
        \label{fig:cluster4_after}
    \end{subfigure}
    \caption{Cluster~4 metrics before and after cluster-specific fine-tuning.}
    \label{fig:cluster4_finetune}
\end{figure}

\textbf{Figure~\ref{fig:cluster4_finetune}} shows the metrics (MSE, RMSE, MAE, $R^2$) before and
after fine-tuning. After removing near-off and quasi-constant turbines, losses decrease
further and $R^2$ rapidly stabilises at a clearly positive level, making Cluster~4 more
consistent with the other clusters. This result suggests that, on top of behaviour-aware
grouping, simple cluster-level rules can effectively reduce residual heterogeneity and
improve forecasting stability without altering the overall federated structure.

\section{Conclusion and Discussion}

\subsection{Key Findings}

\begin{enumerate}
  \item \textbf{Behaviour-aware clustering identifies meaningful turbine behaviors.}
  The proposed federated KMeans with recursive splitting (DRS-auto) partitions 400 turbines into behaviourally coherent groups (high-production, mid-risk, stable, etc.) and isolates a fault/shutdown cluster (two near-offline units) useful for anomaly detection.

  \item \textbf{DRS produces more balanced clusters than k-means++.}
  Both initialisations yield nearly identical forecasting accuracy, but DRS tends to split large clusters more evenly, producing a more practitioner-friendly partition.

  \item \textbf{Privacy-friendly behaviour-aware clustering (DRS).}
    With DRS, raw turbine time series (and per-sample distances) never leave the clients. The server only uses
    \emph{aggregated} $D^2$ scores for roulette-based centre initialisation and then aggregates sample-weighted
    cluster centres and counts, enabling behaviour-aware grouping without centralising turbine-level data.
\end{enumerate}

\subsection{Practical Implications and Limitations}

\paragraph{For system operators and prosumers.}
DSOs, aggregators, and VPPs benefit from high-quality forecasts without centralising raw data. Individual turbine owners can use forecasts for revenue planning and as a ``behavioural benchmark'' to detect emerging faults when actual output deviates significantly from predictions under similar conditions.

\paragraph{Periodic re-clustering and concept drift.}
In practical deployment, turbine behaviour is not stationary: seasonal wind regimes, control policy updates,
maintenance events, and component ageing can gradually shift a turbine’s operating pattern. To keep the grouping
structure aligned with such concept drift, the proposed clustering stage can be re-run on a rolling basis (e.g.,
once every four months). In each cycle, turbines locally recompute the same behavioural summary statistics from
the most recent time window, and the server re-derives an updated cluster tree using the federated auto-split
procedure without collecting raw time series. 

\paragraph{Limitations.}
Current experiments use a relatively clean dataset and do not fully model real-world complexities such as client drop-out, communication failures, and variable data quality. The DRS scheme also tends to produce imbalanced cluster sizes (one or two large ``normal'' clusters plus small anomalous ones), which may affect federated training dynamics.

\clearpage
\appendix
\section{Appendices}
\label{sec:appendix}

\subsection{DRS Algorithm Pseudocode and Parameters}
\label{subsec:app_drs}
The detailed procedure is presented in Table \ref{tab:notation of algorithm},  Algorithm~\ref{alg:drs}.

\begin{table}[!htbp]
\centering\scriptsize
\begin{tabularx}{\textwidth}{|l|X|}
\hline
\textbf{Symbol / Term} & \textbf{Definition} \\
\hline
$C_j$ & Local cluster centers on client $j$ \\
$S_j$ & Number of samples assigned to each cluster on client $j$ \\
$C_k^{\text{global}}$ & Global cluster center $k$ after aggregation \\
$k_{\text{local}}$ & Number of clusters per client (local KMeans) \\
$k_{\text{global}}$ & Number of global clusters \\
$T_{\text{cluster}}$ & Number of communication rounds in the clustering phase \\
$T_{\text{pred}}$ & Number of communication rounds in the prediction phase \\
$f_j^k$ & Local prediction model of client $j$ within cluster $k$ \\
$f_{\text{global}}^k$ & Aggregated global model for cluster $k$ \\
$n_j$ & Number of training samples on client $j$ (FedAvg weight) \\
DRS & Double Roulette Selection — a probabilistic global initialization method \\
D$^2$ Sampling & Distance-based squared sampling for point selection \\
$x_{j,i}$ & The $i$-th data sample on client $j$, used in distance-based sampling \\
FedAvg & Federated Averaging algorithm for model aggregation \\
\hline
\end{tabularx}
\caption{\footnotesize Notation used in the proposed federated clustering and prediction framework}
\label{tab:notation of algorithm}
\end{table}

\begin{algorithm}[H]
\footnotesize
\caption{Federated KMeans clustering with DRS initialization and federated prediction}
\label{alg:drs}
\KwData{Local datasets $\mathcal{D}_j$ for each client $j \in \{1, \dots, M\}$, $q$ is the index over all clients}
\KwResult{Cluster-specific models $f_{\text{global}}^k$ for all turbines}

\textbf{Phase 1: Federated Clustering (DRS Initialization)}\;

$C_{\text{global}} \gets \emptyset$ \tcp*{Initialize empty global centers}

Randomly sample one client $j^*$ and one sample $x^*$ from $\mathcal{D}_{j^*}$\;
Add $x^*$ to $C_{\text{global}}$ \tcp*{First global center initialized}

\For{$k = 2$ \KwTo $k_{\text{global}}$}{
    \For{each client $j$}{
        $D_j \gets \sum_i d^2(x_{j,i}, C_{\text{global}})$ \tcp*{Total D² to current centers}
    }
    Sample client $j^*$ with probability $P(j) = \frac{D_j}{\sum_q D_q}$ \tcp*{First-level roulette}
    Sample $x^*$ from $\mathcal{D}_{j^*}$ with probability $\propto d^2(x, C_{\text{global}})$ \tcp*{Second-level roulette}
    Add $x^*$ to $C_{\text{global}}$
}

Broadcast $C_{\text{global}}$ to all clients \tcp*{Used as initial centers for local KMeans}

$C_{\text{global}} \gets$ weighted average of all $C_j$ using $S_j$\ \tcp*{Initial global centers}

\For{$t = 1$ \KwTo $T_{\text{cluster}}$}{
  Server sends $C_{\text{global}}$ to all clients\ \tcp*{Global update}
  \For{each client $j$}{
    $C_j \gets$ LocalKMeans($\mathcal{D}_j$, init=$C_{\text{global}}$, iter=1)\ \tcp*{One local update}
    $S_j \gets$ updated cluster sizes\;
    Send $(C_j, S_j)$ to server\ \tcp*{Upload updated centers}
  }
  $C_{\text{global}} \gets$ aggregated update from all clients\ \tcp*{Update global centers}
}

\For{each turbine $i$}{
  $x_i \gets$ extract behavioral features\;
  Assign cluster: $\arg\min_k \|x_i - C_k^{\text{global}}\|$\
  \tcp*{\parbox[t]{6cm}{Assign each turbine to its nearest global cluster center}}
}

\vspace{1mm}
\textbf{Phase 2: Federated Prediction}\;

\For{each cluster $k$}{
  Initialize model $f^k_{\text{global}}$\ \tcp*{Model initialization}
  \For{$t = 1$ \KwTo $T_{\text{pred}}$}{
    \For{each client $j$ in cluster $k$}{
      $f_j^k \gets$ LocalTrain($f^k_{\text{global}}, \mathcal{D}_j$)\ \tcp*{Local training}
    }
    $f_{\text{global}}^k \gets$ FedAvg($\{f_j^k\}$, weights=$\{n_j\}$)\ \tcp*{Server aggregation}
  }
}

\For{each turbine $i$}{
  Predict 24h power using assigned model $f_{\text{global}}^k$\ \tcp*{Final forecasting}
}
\end{algorithm}

\subsection{A2: Recursive Auto-split Procedure (DRS-auto)}
\label{app:a2_autosplit}

DRS-auto applies recursive auto-splitting on behavioural features: starting from a root node, it explores federated KMeans configurations and splits based on clustering quality and size constraints. See \textbf{Algorithm~\ref{alg:autosplit_bfs}} and \textbf{Table~\ref{tab:autosplit_params}}.

\begin{table}[!htbp]
\centering\scriptsize
\begin{tabularx}{\textwidth}{|l|>{\raggedright\arraybackslash}X|}
\hline
\textbf{Param} & \textbf{Meaning} \\
\hline
$\mathbf{X}$ & Behavioural feature matrix (standardised). \\
$\mathcal{G}$ & Candidate grid $(n,k,c)$: \#clients, \#clusters, \#comm.rounds. \\
$\tau_{\text{sil}}$ & Silhouette threshold for accepting a split. \\
$\tau_{\min}$ & Min cluster ratio: nodes $\le\tau_{\min}|\mathbf{X}|$ become outlier leaves. \\
$\tau_{\text{large}}$ & Oversized ratio: forces split if node $>\tau_{\text{large}}|\mathbf{X}|$. \\
$Q$, $C_{\text{root}}$, $C_{\text{node}}$ & BFS queue, root node, current node. \\
$\mathcal{L}$ & Labels from \textsc{FederatedKMeans}. \\
\hline
\end{tabularx}
\caption{\scriptsize Key parameters for the recursive auto-split procedure.}
\label{tab:autosplit_params}
\end{table}

\begin{algorithm}[!htbp]
\scriptsize
\caption{Recursive Federated Auto-split Clustering (BFS)}
\label{alg:autosplit_bfs}
\DontPrintSemicolon

\KwIn{$\mathbf{X}$: features; $\mathcal{G}$: param grid; $\tau_{\text{sil}}$, $\tau_{\min}$, $\tau_{\text{large}}$: thresholds}
\KwOut{Hierarchical cluster tree (leaf clusters)}

\textbf{Init:} $C_{\text{root}} \leftarrow$ all samples; $Q \leftarrow \{C_{\text{root}}\}$\;
\While{$Q \neq \emptyset$}{
    $C_{\text{node}} \leftarrow \textsc{Dequeue}(Q)$; $\mathbf{X}_{\text{sub}} \leftarrow$ samples in $C_{\text{node}}$\;
    \If{$|C_{\text{node}}| \le \tau_{\min}|\mathbf{X}|$}{mark as outlier leaf\;}
    \Else{
        $s^*, \mathcal{L}^*, p^* \leftarrow -\infty, \texttt{None}, \texttt{None}$\;
        \ForEach{$(n,k,c)\in \mathcal{G}$}{
            $\mathcal{L} \leftarrow \textsc{FedKMeans}(\mathbf{X}_{\text{sub}}; n,k,c)$\;
            \If{$|\mathrm{unique}(\mathcal{L})|>1$}{
                $s \leftarrow \text{silhouette}(\mathbf{X}_{\text{sub}},\mathcal{L})$\;
                \lIf{$s > s^*$}{$s^*, \mathcal{L}^*, p^* \leftarrow s, \mathcal{L}, (n,k,c)$}
            }
        }
        $\textit{split} \leftarrow (\mathcal{L}^*\neq \texttt{None}) \land (s^*\ge \tau_{\text{sil}} \lor |C_{\text{node}}| > \tau_{\text{large}}|\mathbf{X}|)$\;
        \If{$\neg\textit{split}$}{mark $C_{\text{node}}$ as leaf\;}
        \Else{
            \ForEach{label $\ell$ in $\mathcal{L}^*$}{
                $C_{\text{child}} \leftarrow \{x : \mathcal{L}^*(x)=\ell\}$; add as child of $C_{\text{node}}$\;
                \lIf{$|C_{\text{child}}| > \tau_{\min}|\mathbf{X}|$}{$\textsc{Enqueue}(Q, C_{\text{child}})$}
                \lElse{mark as outlier leaf}
            }
        }
    }
}
\end{algorithm}

\bibliographystyle{plainnat}
\bibliography{references}
\end{document}